\documentclass{isprs} 
\usepackage{subfigure}
\usepackage{amsmath} 
\usepackage{setspace}
\usepackage{geometry} 
\usepackage{epstopdf}
\usepackage[labelsep=period]{caption}  
\usepackage[british]{babel} 
\usepackage[hang]{footmisc}
\usepackage[dvipsnames]{xcolor}
\usepackage{cleveref}
\usepackage{amsthm,amsmath,amssymb}
\usepackage{mathrsfs}

\usepackage{tikz}

\newcommand{\Nerf}{{NeRF}}
\newcommand{\OurNeRF}{{SparseSat-NeRF}} 
\newcommand{\OurNeRFShort}{{SpS-NeRF}} 

\crefformat{section}{\S#2#1#3} 
\crefformat{subsection}{\S#2#1#3}
\crefformat{subsubsection}{\S#2#1#3}

\begin{document}

\title{\OurNeRF: Dense Depth Supervised Neural Radiance Fields for Sparse Satellite Images}

\author{
 Lulin Zhang\textsuperscript{1,2}, Ewelina Rupnik\textsuperscript{2}}

\address{
	\textsuperscript{1 } Université de Paris, Institut de Physique du Globe de Paris, CNRS, Paris, France - lzhang@ipgp.fr\\ 
	\textsuperscript{2 } Université de Gustave Eiffel, IGN-ENSG, LaSTIG, Saint-Mandé, France - ewelina.rupnik@ign.fr\\
}

\commission{II, }{WG II/1} 
\workinggroup{II/1} 
\icwg{}   

\abstract{Digital surface model generation using traditional multi-view stereo matching (MVS) performs poorly over non-Lambertian surfaces, with asynchronous acquisitions, or at discontinuities. Neural radiance fields (\Nerf) offer a new paradigm for reconstructing surface geometries using continuous volumetric representation. \Nerf~is self-supervised, does not require ground truth geometry for training, and provides an elegant way to include in its representation physical parameters about the scene, thus potentially remedying the challenging scenarios where MVS fails. However, \Nerf~and its variants require many views to produce convincing scene's geometries which in earth observation satellite imaging is rare. In this paper we present \OurNeRF~(\OurNeRFShort) -- an extension of Sat-\Nerf~adapted to sparse satellite views. \OurNeRFShort~employs dense depth supervision guided by cross-correlation similarity metric provided by traditional semi-global MVS matching. We demonstrate the effectiveness of our approach on stereo and tri-stereo Pléiades 1B/WorldView-3 images, and compare against \Nerf~and Sat-\Nerf. The code is available at \textcolor{magenta}{\texttt{https://github.com/LulinZhang/SpS-NeRF}}}


\maketitle

\section{Introduction}
Satellite imagery and 3D digital surface models (DSM) derived from them are used in a wide range of applications, including urban planning, environmental monitoring, geology, disaster rapid mapping, etc. Because in many of those applications the quality of the DSMs is essential, a vast amount of research has been undertaked in the last few decades to enhance their precision and fidelity.\\
Classically, DSMs are derived from images with semi-global dense image matching \cite{hirschmuller2005accurate,pierrot2006multiresolution} (SGM) followed by a depth map fusion step \cite{rupnik20183d} or more recently with hybrid \cite{konrad2017} or end-to-end \cite{PSMNet} deep learning based approaches. A new way of solving the dense image correspondence problem is proposed by Neural Radiance Fields (\Nerf) \cite{Mildenhall20eccv_nerf}. Unlike the traditional methods, \Nerf~leverages many views to learn to represent the scene as a continuous volumetric representation (i.e., 3D radiance field). This representation is defined by a neural network and has the unique capacity to incorporate different aspects of the physical scene, e.g., surface radiance or illumination sources.\\
Despite the tremendous \textit{hype} around the neural radiance fields, the \textit{state-of-the-art} results remain conditioned on a rather large number of input images. With few input images, \Nerf~has the tendency to fit incorrect geometries, possibly because it does not know that the majority of the scene is composed of empty space and opaque surfaces. In a space-borne setting, it is rare to have many images of a given scene acquired under multiple viewing angles within a defined time window. With the exception of the Pléaides \textit{persistent surveillance} collection mode, the most common configuration includes a stereo pair or a stereo-triplet of images. Previous works have attempted to apply \Nerf~on satellite images, including S-\Nerf~\cite{derksen2021shadow} and Sat-\Nerf~\cite{mari2022sat}, but they bypassed the problem of sparse input views by using multi-date images. 

\begin{figure}[htbp]
	\begin{center}

		\subfigure[Input]{
			\begin{minipage}[t]{0.19\linewidth}
				\centering
				\includegraphics[width=1\linewidth]{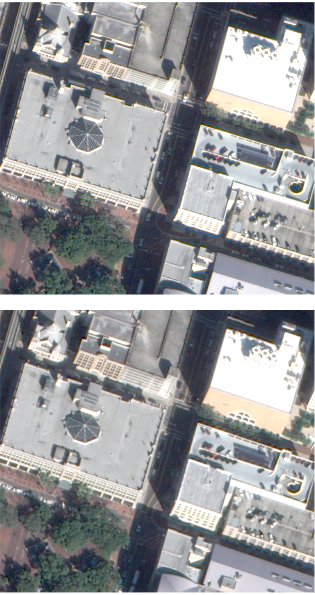}
			\end{minipage}%
		}
		\subfigure[\Nerf]{
			\begin{minipage}[t]{0.18\linewidth}
				\centering
                    \begin{tikzpicture}
                    \node[anchor=south west,inner sep=0] (image) at (0,0) {
				\includegraphics[width=1\linewidth]{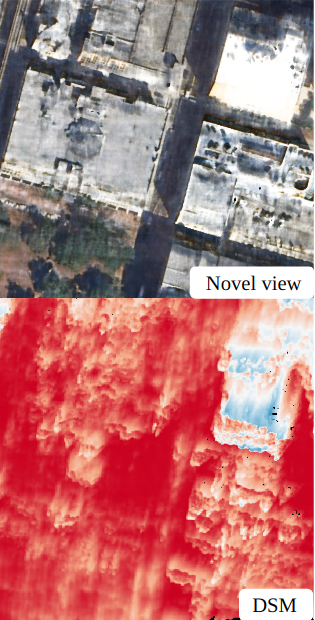}};
                    \begin{scope}[x={(image.south east)},y={(image.north west)}]
                    \draw[red,thick] (0,0.52) rectangle (1,1);
                    \draw[blue,thick] (0,0) rectangle (1,0.515);
                    \end{scope}
                    \end{tikzpicture}
			\end{minipage}%
		}
		\subfigure[Sat-\Nerf]{
			\begin{minipage}[t]{0.18\linewidth}
				\centering
                    \begin{tikzpicture}
                    \node[anchor=south west,inner sep=0] (image) at (0,0) {
				\includegraphics[width=1\linewidth]{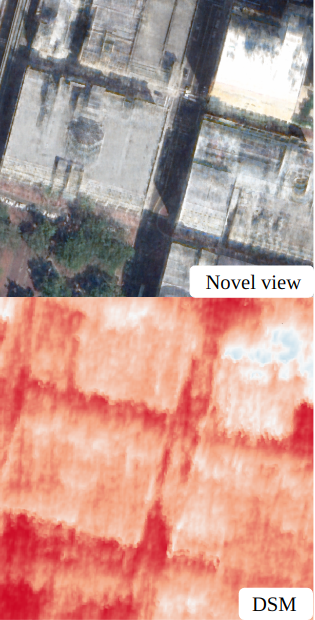}};
                    \begin{scope}[x={(image.south east)},y={(image.north west)}]
                    \draw[red,thick] (0,0.52) rectangle (1,1);
                    \draw[blue,thick] (0,0) rectangle (1,0.515);
                    \end{scope}
                    \end{tikzpicture}    
			\end{minipage}%
		}
		\subfigure[Ours]{
			\begin{minipage}[t]{0.18\linewidth}
				\centering
                    \begin{tikzpicture}
                    \node[anchor=south west,inner sep=0] (image) at (0,0) {
				\includegraphics[width=1\linewidth]{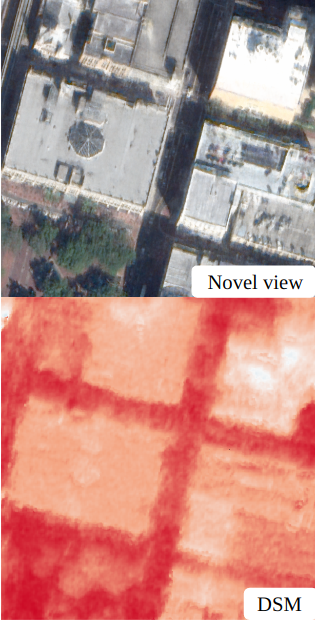}};
                    \begin{scope}[x={(image.south east)},y={(image.north west)}]
                    \draw[red,thick] (0,0.52) rectangle (1,1);
                    \draw[blue,thick] (0,0) rectangle (1,0.515);
                    \end{scope}
                    \end{tikzpicture}        
			\end{minipage}%
		}
		\subfigure[GT]{
			\begin{minipage}[t]{0.178\linewidth}
				\centering
                    \begin{tikzpicture}
                    \node[anchor=south west,inner sep=0] (image) at (0,0) {
				\includegraphics[width=1\linewidth]{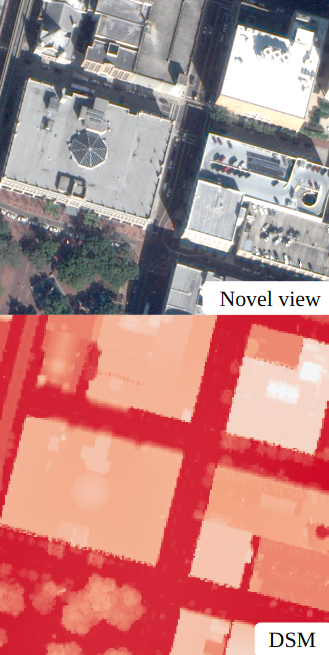}};
                    \begin{scope}[x={(image.south east)},y={(image.north west)}]
                    \draw[red,thick] (0,0.52) rectangle (1,1);
                    \draw[blue,thick] (0,0) rectangle (1,0.515);
                    \end{scope}
                    \end{tikzpicture}         
			\end{minipage}%
		}
            \caption{\textbf{\OurNeRFShort} (Ours) \textbf{and competitors}. {\Nerf~variants trained on 2 images. Our network leverages dense depth information calculated by stereo-matching on downsampled images. Compared to \Nerf~and Sat-\Nerf, \OurNeRFShort~renders sharper novel views (\textcolor{red}{$\square$}), reconstructs more reliable 3D geometries (\textcolor{blue}{$\square $}).}} 
		\label{teaser}
	\end{center}
\end{figure}
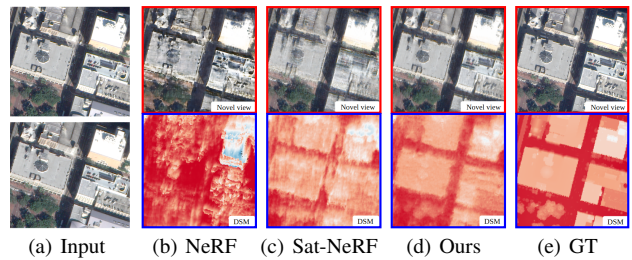

\paragraph{Contributions.}
In this paper, we present a \Nerf~variant that attains \textit{state-of-the-art} results in novel view synthesis and 3D reconstruction using sparse single-date satellite images. Inspired by the architecture proposed in \cite{mari2022sat}, we lay down its extension adapted to sparse satellite views and refer to it as \OurNeRF, or \OurNeRFShort. Precisely 
 \begin{itemize}
     \item we adopt low resolution dense depths generated with traditional MVS for supervision and consequently enable the generation of novel views and 3D surfaces from sparse satellite images. We demonstrate the efficiency of this method on as few as two and three input views;
     \item we increase the robustness of the predicted views and surfaces by incorporating correlation-based uncertainty into the guidance of NeRF using depth information;
     \item we provide in-depth analysis of the benefits of adding dense depth supervision into the NeRF architecture. 
 \end{itemize}


\section{Related work}
\paragraph{Image matching vs \Nerf}
Traditional stereo or multi-view stereo (MVS) matching approaches \cite{hirschmuller2005accurate,gallup2007real,bleyer2011patchmatch,bulatov2011multi,furukawa2009accurate} establish correspondences between pixels in different images by calculating patch-based similarity metrics such as correlation coefficient or mutual information. Although these methods often produce impressive results in favourable matching conditions, they tend to struggle with images lacking texture, at discontinuities or in the presence of non-Lambertian surfaces such as forest canopies or icy surfaces.
Learning-based MVS methods \cite{bittner2019late,stucker2020resdepth,gao2021rational,gomez2022experimental,huang2018deepmvs}  
 attempt and often succeed in overcoming those challenges, however, they require very precise and up-to-date ground truth depth maps for training and those are difficult to obtain in a satellite setting.  In contrast, \Nerf~offers a self-supervised deep learning approach without resorting to ground truth geometry, and relying exclusively on images at input. Because it operates on a truly single-pixel level, it overcomes the shortcomings of traditional patch-based methods \cite{buades2015reliable}. Furthermore, \Nerf~defined as a function of radiance accumulated along image rays opens up the possibility to model physical parameters of the scene such as reflectance of scene's materials.

\paragraph{\Nerf~variants towards fewer input views.}
Vanilla \Nerf~relies exclusively on RGB values to maintain consistency between training images. Consequently, it requires a large number of images to resolve the ambiguity embedded within the modelled volumetric fields. This greediness of \Nerf~has been addressed across several research works, which focus on adding priors through incorporating semantic information, or sparse/ dense depth supervision. The latter is particularly interesting because \textit{Structure from Motion} (SfM) or the subsequent MVS matching provide reliable depth information. Additionally, in satellite imaging, the dense depth information is available without extra processing through, e.g., the global SRTM elevation model.

\paragraph{Learning priors with semantics.}
{PixelNeRF} demonstrates excellent results in novel view synthesis over an unknown scene with only one view. To this end, \cite{yu2021pixelnerf}  extend the canonical \Nerf~with deep features and pre-train the entire architecture enabling its generalization to new scenes. Analogously, Diet\Nerf~\cite{jain2021putting} adopts a pre-trained visual transformer (ViT) and enforces consistent semantics across all views (including the novel view). Sin\Nerf~\cite{xu2022sinnerf} extends further this idea by combining global semantics using the self-supervised Dino ViT, then instead of using image feature embeddings leverages the classification token representation, thus making their approach less susceptible to pixel misalignments between views. Sin\Nerf~also employs local texture regularization and depth supervision through depth warping to novel views. {MVSNeRF}  \cite{chen2021mvsnerf} borrows from multi-view stereo matching in projecting 2D convolutional neural networks (CNN) features to planes sweeping through the scene. 3D CNNs are then used to extract a neural encoding volume, which once regressed translate to RGB and density. 
%

\paragraph{Sparse depth supervision} DS-NeRF \cite{deng2022depth} was the first to propose sparse depth supervision using 3D points obtained from SfM. The authors propose an adapted ray sampling strategy and a depth termination loss weighted by the 3D point's reprojection error.  Sat-NeRF \cite{mari2022sat} applied the same sparse depth supervision in multi-date satellite images, reducing the number of training images to approximately $15$. Interestingly, Sat-\Nerf~architecture includes scene's physical parameters specific to earth observation satellites such as albedo and solar correction (for asynchronous acquisitions).

\paragraph{Dense depth supervision.} NerfingMVS \cite{wei2021nerfingmvs} combines learning-based multi-view stereo with \Nerf~for indoor mapping. Starting from a set of sparse 3D points output from SfM, NerfingMVS first trains a monocular dense depth prediction network. Consistency checks between per-view predicted depths serve as error maps and guide the following ray sampling in the final \Nerf~optimization. In their most view-sparse scenario 35 images are available for training. 
Similarily, Roessle et al. \cite{roessle2022dense} (referred to in the following as DDP\Nerf) incorporate dense depth supervision in their \Nerf~variant. However, unlike in NerfingMVS where dense depths are guessed from single views, DDP\Nerf~learns a depth completion network from sparse depth maps. This, together with an explicit depth loss, makes it a better performing method. Experiments demonstrate good performance with as few as 18 train images.

The above methods resort to learning-based dense depth prediction because their focus is on indoor scenes, with textureless surfaces where traditional MVS might fail. In our real world satellite scenario this is, in general, less of an issue and we demonstrate that dense image matching with SGM is good enough to guide the \Nerf~optimization.

\section{Methodology}
\begin{figure*}[htbp]
	\begin{center}
				\centering
				\includegraphics[trim = 10 150 60 50, clip, width=0.8\linewidth]{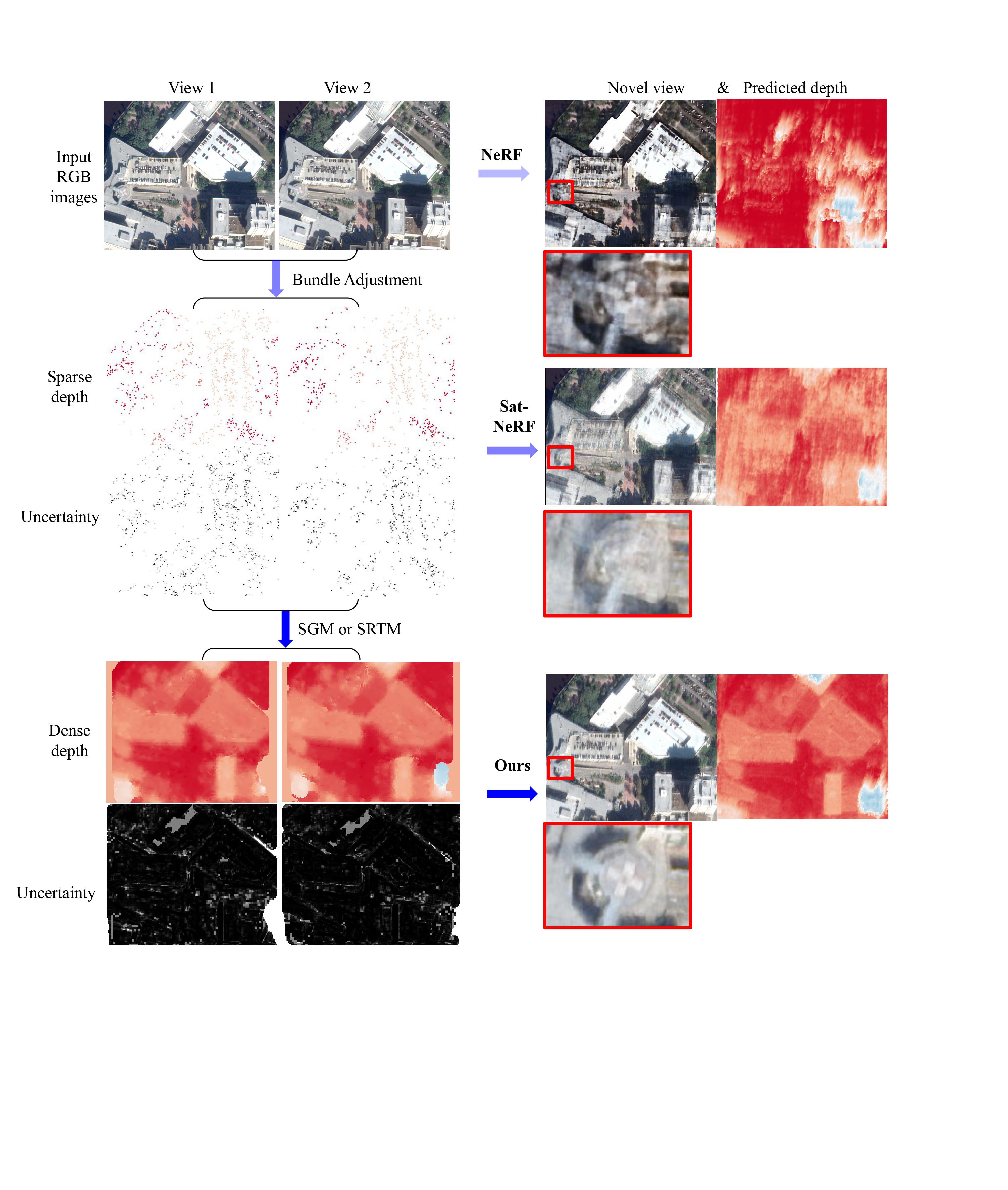}
    \caption{\textbf{Workflows of \OurNeRFShort~(Ours), Sat-\Nerf} \textbf{and \Nerf}. In our experimental setting we use 2 or 3 satellite images to optimize the neural radiance fields for photo-realistic novel view rendering, and for DSM recovery. Without any depth supervision, NeRF fails to render high quality novel views and DSM. Sat-NeRF incorporates sparse depth information and uses the bundle adjustment re-projection errors as uncertainties to weigh the depth loss; it improves the results, but the artifact remain present due to the insufficient number of training views. \OurNeRFShort~further employs low resolution dense depth maps from traditional methods such as SGM, and uses the $\left(1 - correlation \right)$ score as uncertainty, and takes advantage of the dense depth to guide sampling along the casted ray, leading to improved performance.}
    \label{workflow}
	\end{center}
\end{figure*}

Our method builds on top of Sat-\Nerf~\cite{mari2022sat} and DDP\Nerf~\cite{roessle2022dense}. We borrow from Sat-\Nerf~the general architecture save for the transient objects and solar correction modelling as we deal with synchronous acquisitions. We add a dense depth supervision and depth loss similar to the one proposed in DDP\Nerf, but we replace the depth loss distance metric and define an uncertainty based on SGM's correlation maps. The workflows of \Nerf, Sat-\Nerf~and \OurNeRFShort~are illustrated in \Cref{workflow}.

\subsection{Neural Radiance Fields Preliminaries}
\Nerf~\cite{Mildenhall20eccv_nerf} learns a continuous volumetric representation of the scene from a set of images characterised by the sensor position and the viewing direction. This representation is defined by a fully-connected (non-convolutional) deep network. 
It samples $N$ query points along each camera ray through the 3D field and integrate the weighted radiance to render each pixel, and optimize the NeRF network $F_{\Theta}$ by imposing the rendered pixel values to be close to the training images. 
For each query point, NeRF simultaneously models the volume density $\sigma$ and the emitted radiance $\textbf{c} = (r, g, b)$ at that 3D point $\textbf{x} = (x, y, z)$ from the viewing angle $\textbf{d} = (d_x, d_y, d_z)$:\\
\begin{equation}
F_{\Theta}(\textbf{x}, \textbf{d}) = (\textbf{c}, \sigma)~.  
\end{equation}
Each camera ray $\textbf{r}$ is defined by a point of origin $\textbf{o}$ and a direction vector $\textbf{d}$ as $\textbf{r}(t) = \textbf{o} + t\textbf{d}$. Each query point in $\textbf{r}$ is defined as $\textbf{x}_i = \textbf{o} + t_i\textbf{d}$, where $t_i$ locates between the near and far bounds of the scene, $t_n$ and $t_f$. The rendered pixel value $\textbf{C}(\textbf{r})$ of ray $\textbf{r}$ is calculated as:
\begin{gather}
    \textbf{C(r)} = \sum_{i=1}^{N} {T_i {\alpha}_i c_i}~, \nonumber\\[1ex]
    {\alpha}_i = 1 - e^{-{\sigma}_i {\delta}_i}~,\quad 
    T_i = \prod_{j=1}^{i-1} ({1 - {\alpha}_j})~, \quad 
    {\delta}_i = t_{i+1} - t_i~,
    \label{NeRFCr}
\end{gather}
where $\alpha_i$ represents the opacity of the current query point $\textbf{x}_i$, $T_i$ stands for the probability that $\textbf{x}_i$ reaches the ray origin $\textbf{o}$ without being blocked. In other words, the color $c_i$ of the current query point $\textbf{x}_i$ contributes to the accumulated color $\textbf{C}(\textbf{r})$ only if it is highly opaque (i.e., large value of $\alpha_i$) and there are no opaque particles in front of it (i.e., high value of $T_i$).


\subsection{\OurNeRF} 
\paragraph{Pre-processing.}
Following the Sat-\Nerf's pipeline, the RPC-poses of our input images are first refined in a bundle adjustment. Then, for $N$ input images, we run $N$ independent SGMs to obtain a low-resolution depth map for each image (i.e., scale factor of $2^{-2}$). We choose to rely on low resolution depths to (i) avoid biasing our \OurNeRFShort~towards the SGM solution; and (ii) because high resolution depths might provide incomplete depth information at challenging surfaces (e.g., low texture). The depth maps are accompanied by similarity metrics that will further act as depth prediction quality measures in supervising the \OurNeRFShort. In our case, the metric is the cross-correlation map. If low-resolution depth maps are not available, the SGM depths can be replaced by coarse global DEM such as SRTM (with the similarity metric globally set to a constant value).

\paragraph{Depth supervision.} 
Our goal is to include the depth prior in \OurNeRFShort~optimization. Analogously to the formulation presented in \cite{roessle2022dense}, three ingredients are necessary for that end: (i)~a way to predict the depth of a given ray by accumulating radiance fields throughout the optimized volume; (ii)~description of the sample distribution along a given ray; and finally (iii)~input depth maps and their associated uncertainty metrics.
The depth prediction along a ray $D$(\textbf{r}) can be calculated as:
\begin{equation}
    {D}(\textbf{r}) = \sum_{i=1}^{N} {T_i {\alpha}_i t_i}~,
    \label{d_r}
\end{equation}
where the depth $t_i$ of the current sample point $i$  would contribute to the accumulated depth  ${D}(\textbf{r})$ if it is opaque, ignoring the sample points in front of $t_i$.
To characterise the samples' distribution along the ray we follow the standard deviation equation \cite{roessle2022dense}: 
\begin{equation}
    S(\textbf{r})^2 = \sum_{i=1}^{N} {T_i {\alpha}_i (t_i - D(\textbf{r}))^2}~.
    \label{d_S}
\end{equation}
%
Here, lower standard deviation values indicate samples located around the estimated depths and lead to sharper edges at object surfaces. We now define an equivalent uncertainty driven by our input data, i.e., the similarity metrics produced by SGM:
\begin{equation}
    \Sigma(\textbf{r}) = \gamma \cdot (1 - \text{corr}(\textbf{r})) + m~,
    \label{s_GT}
\end{equation}
where $\text{corr}(\textbf{r})$ is the cross-correlation similarity for a ray sample at the input depth, $\gamma$ and $m$ are the normalizing scaling and shift parameters, in our experiments empirically set to $1.0$ and $10e^{-4}$. The uncertainty measure (\Cref{s_GT}) intervenes three times during the optimization: (i) as a weight applied to the final depth loss; (ii) as a threshold to determine whether the loss should be activated; and (iii) in guided ray sampling (see next paragraph). 

All ingredients combined constitute the depth loss encouraging depths' predictions $D(\textbf{r})$ to be close to the input dense depths $ \overline{D}(\textbf{r})$, guided by the input uncertainty:
\begin{equation}
    \mathcal{L}_{depth}(\textbf{r}) = \sum_{\textbf{r} \in R_{sub}} (\text{corr}(\textbf{r})(D(\textbf{r}) - \overline{D}(\textbf{r}))^2~.
    \label{depthloss}
\end{equation}
The $R_{sub}$ is defined as a ray's subregion where either of the two conditions are satisfied: (1) $S(\mathbf{r}) > \Sigma(\mathbf{r})$; (2) $\left|(D(\textbf{r}) - \overline{D}\textbf{r})\right| > \Sigma(\mathbf{r})$. Those bounds favour ray termination within  $\left(1\cdot\Sigma\right)$ from our depth priors \cite{roessle2022dense}. Outside this region, the depth loss is inactive or clipped. The depth loss participates in all training iterations.
\paragraph{Total loss.}
Our \OurNeRFShort~is supervised with the ground truth pixel color $\textbf{$\overline{\mathbf{C}}$}(\textbf{r})$  and the dense depth information $\overline{D}$(\textbf{r}) weighted by the quality metric corr(\textbf{r}). Following \Cref{NeRFCr}, the color (RGB) of a pixel is rendered through the accumulation of the RGB values of samples along the casted ray. The color loss encourages the predicted pixel colors $\textbf{C}(\textbf{r})$ to be as close as possible to the ground truth colors and is defined on a set $R$ containing all ray samples (there is no clipping unlike in the depth loss):
\begin{equation}
    \mathcal{L}_{color}(\textbf{r}) = \sum_{\textbf{r} \in R} \left \| \textbf{C}(\textbf{r}) - \textbf{$\overline{\mathbf{C}}$}(\textbf{r}) \right \| _2 ^2~.
    \label{colorloss}
\end{equation}
The \OurNeRFShort's~total loss is thus a combination of \Cref{colorloss} and \Cref{depthloss}:
\begin{equation}
    \mathcal{L} = \mathcal{L}_{color}(\textbf{r}) + \lambda \mathcal{L}_{depth}(\textbf{r})~,
    \label{loss}
\end{equation}
where $\lambda$ is a weight balancing the color and depth contributions. We empirically found that $\lambda = \frac{1}{3}$ performs best in urban areas and $\lambda = \frac{50}{3}$ in rural areas.
 
\begin{figure*}[htbp]
	\begin{center}
		\subfigure[A selected image row (\textcolor{magenta}{\textbf{--}})]{
			\begin{minipage}[t]{0.280\linewidth}
				\centering
				\includegraphics[width=0.8\linewidth]{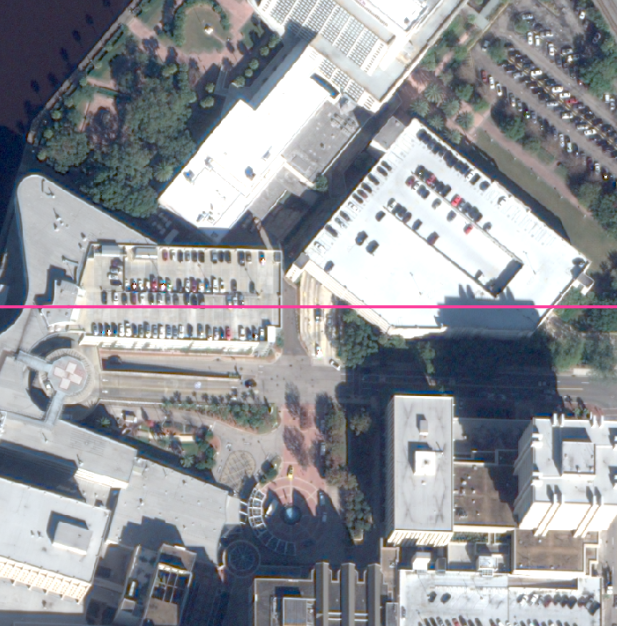}
			\end{minipage}%
		}
		\subfigure[Ray sampling along the selected image row]{
			\begin{minipage}[t]{0.575\linewidth}
				\centering
				\includegraphics[width=0.8\linewidth]{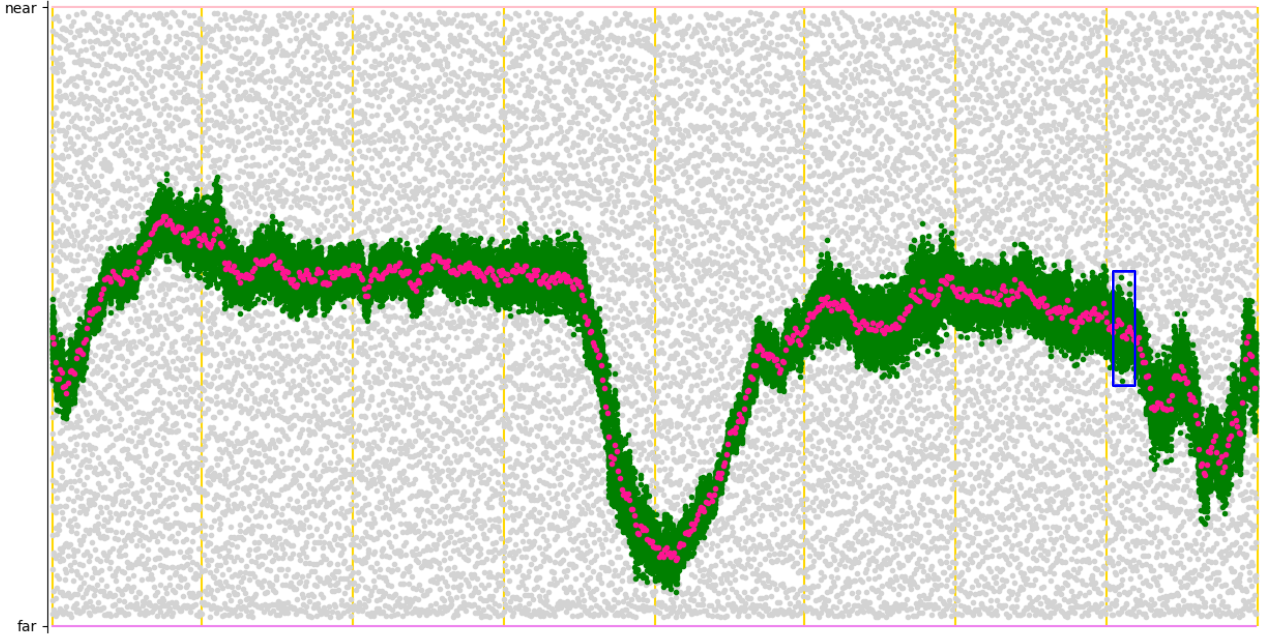}
			\end{minipage}%
		}      
		\subfigure[Zoom]{
			\begin{minipage}[t]{0.06\linewidth}
				\centering
				\includegraphics[width=0.71\linewidth]{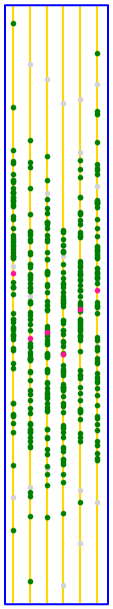}
			\end{minipage}%
		}      
		\caption{\textbf{Ray sampling}. The samples in (b) correspond to the selected image row in (a), while in (c) we zoom over a few ray samples. Similarily to Roessle et al., we divide ray samples in two groups of the same cardinality (i.e., $2\times64$). The first group draws samples (\textcolor{Gray}{\textbf{- -}}) within the near and far planes. At inference, the second group draws samples (\textcolor{ForestGreen}{\textbf{- -}}) following a Gaussian distribution around the estimated dense depths $D(\textbf{r})$ (\textcolor{magenta}{\textbf{- -}}) (see \Cref{d_r}), their upper and lower bounds are defined by the estimated standard deviation $S(\textbf{r})$ (see \Cref{d_S}). At train time we use the input depths and their corresponding uncertainties $\{\overline{D},\Sigma\}$. The yellow lines (\textcolor{YellowOrange}{{$\mathbf{|}$}}) represent the rays.}
		\label{raysample}
	\end{center}
\end{figure*}

\paragraph{Ray sampling}\label{para:sample} We adopt guided sampling from~\cite{roessle2022dense}, whose approach takes advantage of depth cues to efficiently query samples. 
It substitutes the hierarchical sampling coarse network in the original \Nerf. More specifically, the ray samples are divided into two groups queried sequentially. The  points of the first group are sampled randomly within the entire scene's envelope, while the second group of points is concentrated around the known input (train) or predicted (test) surface. The points around the surface are spread following a Gaussian distribution determined by (1) the input depth $N(\overline{D}(\textbf{r})$, $\Sigma(\textbf{r}))$ for the pixels with input depth information during training; or (2) the estimated depth $N(D(\textbf{r})$, $S(\textbf{r}))$ for all the pixels during testing, as well as the pixels without input depth during training (e.g., SGM provides no depth in occluded areas). We illustrated the distribution of the rays sampled by this strategy in \Cref{raysample}.
%


 
\section{Experiments}
\begin{figure}[t!]
	\begin{center}
				\centering
				\includegraphics[width=1\linewidth]{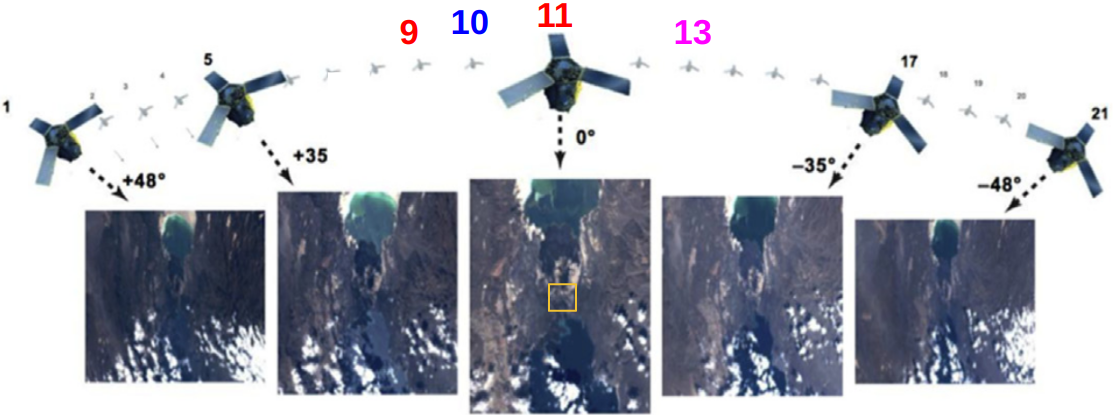}
		\caption{\textbf{Djibouti dataset}. The images labeled $\{{\textcolor{red}{\textbf{9}, \textbf{11}}}\}$ are used for training 2-views scenario, and the images labeled $\{{\textcolor{red}{\textbf{9}, \textbf{11}}}, \textcolor{magenta}{\textbf{13}}\}$ are used for training 3-views scenario. The image labelled $\{\textcolor{blue}{\textbf{10}}\}$ is used for testing both scenarios. The remaining images are ignored. The yellow rectangle (\textcolor{yellow}{\textbf{$\square$}}) represents the area of interest cropped for our experiments.}\label{DjiboutiFlight}
	\end{center}
\end{figure}

We conduct experiments on two datasets:
\begin{itemize}
    \item \textbf{Djibouti dataset} located in the Asal-Ghoubbet rift, Republic of Djibouti, introduced in \cite{labarre2019retrieving} and illustrated in \Cref{DjiboutiFlight}. It represents a series of 21 multiangular Pléiades images collected in a single ﬂyby on January 26, 2013. During training we use only two or three RGB cropped images ($\sim$ 800 $\times$ 800 px), with 2m Ground Sampling Distance (GSD). 
    \item  \textbf{DFC2019 dataset} The 2019 IEEE GRSS Data Fusion Contest \cite{le20192019} contains different areas of interest (AOI) in the city of Jacksonville, Florida, USA, prov-iding in total 26 WorldView-3 images collected between 2014 and 2016. We choose the AOI 214 as it contains 3 images taken at the same time and use it to train two independent networks: with 2 and 3 views used in the training images. For novel view generation, we choose another image from the dataset and consider it the ground truth. Because \OurNeRFShort~does not model transient objects, our goal was to minimize the acquisition time gap and respect the seasonality in choosing the novel views. The sun elevation, azimuth and the acquisition time of the 4 selected images are displayed in the table. 
\end{itemize}


\begin{table}[t!]
\centering
\begin{tabular}{|c|c|c|c|}
\hline
\textbf{Image} & \textbf{Sun} & \textbf{Sun} & \textbf{Acquisition} \\ 
\textbf{name}   & \textbf{elevation} & \textbf{azimuth} & \textbf{date}  y-m-d\\\hline\hline
007 & 33.5 & 158.9 & 14-12-27$^{16:11:09}$ \\ \hline 
008 & 36 & 155.0 & 15-01-21$^{16:12:43}$ \\ \hline
009 & 36 & 155.1 & 15-01-21$^{16:12:53}$ \\ \hline
010 & 36 & 155.2 & 15-01-21$^{16:13:08}$ \\ \hline
\end{tabular}
\caption{\textbf{DFC2019 dataset, AOI 214}. During training, we use the following subsets of images: $\{ 009, 010 \}$, $\{ 008, 009, 010 \}$.}
\label{DFC2019}
\end{table}

\subsection{Implementation details}
We use Sat-\Nerf~as the backbone architecture (lr=$1e^{-5}$, decay= $0.9$, batch\_size=$1024$). Our focus is on sparse views captured synchronously from the same orbit thus we disable the uncertainty weighting for transient objects and the solar correction. 
We also disable the two components for Sat-\Nerf~because our experiments are conducted on single-epoch images. In contrast to \Nerf~and Sat-\Nerf, \OurNeRFShort~uses only the coarse architecture (no fine model) with 64 initial samples and 64 guided samples (\textcolor{Gray}{\textbf{- -}} and \textcolor{ForestGreen}{\textbf{- -}} in \Cref{raysample}). For a fair comparison the number of samples and \textit{importance} samples (i.e., fine model) in \Nerf~and Sat-\Nerf~are also 64 each. We optimize \OurNeRFShort~for 30k iterations, which takes $\sim$2 hours on NVIDIA GPU with 40GB RAM. The input low resolution DSMs were computed from images downscaled by a factor of 4 \\
({$SGM_{scl4}$}).


\subsection{Evaluation}
Tests are carried out using 2 and 3 views leading to 4 scenarios:
\begin{enumerate}
    \item $DFC_{2v}$, test on 008 and train on $\{009, 010\}$;
    \item $DFC_{3v}$, test on 007 and train on $\{008, 009, 010\}$;
    \item $Dji_{2v}$, {test on 10 and train on $\{9, 11\}$;}
    \item $Dji_{3v}$, {test on 10 and train on $\{9, 11, 13\}$.} 
\end{enumerate}
    
We evaluate the performance of \OurNeRFShort~qualitatively and quantitatively on 2 tasks: (1) novel view synthesis and (2) altitude extraction. Precision metrics are Peak Signal-to-Noise Ratio (PSNR) and Structural Similarity Index measure (SSIM) \cite{wang2004image} for view synthesis, and Mean Altitude Error (MAE) for altitude extraction. We differentiate between MAE$_{in}$ and MAE$_{out}$ for errors computed on valid pixels and invalid pixels (e.g., due to low correlation or occlusions). The classification into valid and invalid pixels is produced by SGM. Ground truth (GT) images are \textit{true} images not seen during training, while GT DSMs are a LiDAR acquisition for the DFC2019 dataset, and a photogrammetric DSM generated with 21 high-resolution panchromatic Pléiades images (GSD=$0.5m$) for Djibouti dataset. \OurNeRFShort~is also compared with competitive vanilla \Nerf, Sat-\Nerf, {and DSMs generated with SGM using full-resolution images (i.e., $SGM_{scl1}$)}. 


\begin{table*}[htbp]
\scriptsize
\centering
\begin{tabular}{|l|c|c|c|c||c|c|c|c||c|c|c|c||c|c||}\hline
& \multicolumn{4}{c||}{PSNR $\uparrow$} & \multicolumn{4}{c||}{SSIM $\uparrow$} & \multicolumn{4}{c||}{MAE$_{in}$ $\downarrow$} & \multicolumn{2}{c||}{MAE$_{out}$ $\downarrow$}\\\hline
& DFC$_{2v}$ & DFC$_{3v}$ & Dji$_{2v}$ & Dji$_{3v}$ & DFC$_{2v}$ & DFC$_{3v}$ & Dji$_{2v}$ & Dji$_{3v}$ & DFC$_{2v}$ & DFC$_{3v}$ & Dji$_{2v}$ & Dji$_{3v}$& DFC$_{2v}$ & DFC$_{3v}$ \\\hline\hline
NeRF & 12.89 & 14.56 & 27.8 & 35.22 & 0.65 & 0.67 & 0.8 & 0.94 & 9.51 & 6.56 & 9.72 & 14.44  & 13.2 & 11.98\\\hline
Sat-NeRF & 17.72 & 18.46 & 32.3 & 36.17 & 0.8 & 0.83 & 0.9 & \textbf{0.95} & 5.89 & 4.63 & 9.51 & 10.11  & 11.75 & 7.53\\\hline
\OurNeRFShort & \textbf{20.2} & \textbf{19.06} &\textbf{ 32.85} & \textbf{36.26} & \textbf{0.87} & \textbf{0.86} & \textbf{0.92} & \textbf{0.95}  & \textcolor{magenta}{3.02} & \textcolor{magenta}{2.86} & \textcolor{magenta}{1.57} & \textcolor{magenta}{1.35}  & \textcolor{blue}{7.77} & \textcolor{blue}{5.62} \\\hline
$SGM_{scl1}$ & / & / & / & / & / & / & / & / & \textcolor{blue}{2.77} & \textcolor{blue}{2.05} & \textcolor{blue}{1.15} & \textcolor{blue}{0.81}  & \textcolor{magenta}{9.82} & \textcolor{magenta}{6.68} \\\hline
\end{tabular}
\caption{\textbf{Quantitative metrics}. Best performing metrics in PSNR and SSIM are in bold, while best and second best performing metrics in MAE$_{in}$ and MAE$_{out}$ are in \textcolor{blue}{blue} and \textcolor{magenta}{magenta}. \OurNeRFShort~outperformed \Nerf~and Sat-\Nerf~in all the scenarios. \OurNeRFShort~is less good than SGM$_{scl1}$ in altitude extraction on valid pixels (MAE$_{in}$) which we attribute to the lack of regularization. However, \OurNeRFShort~is better than SGM$_{scl1}$ in occluded and poorly textured areas (MAE$_{out}$). Note that no invalid pixels were identified for the Djibouti dataset.  
}
\label{PSNR_SSIM_MAE}
\end{table*}

\begin{figure*}[htbp]
	\begin{center}

		\subfigure[NeRF $DFC_{2v}$]{
			\begin{minipage}[t]{0.19\linewidth}
				\centering
				\includegraphics[width=1\linewidth]{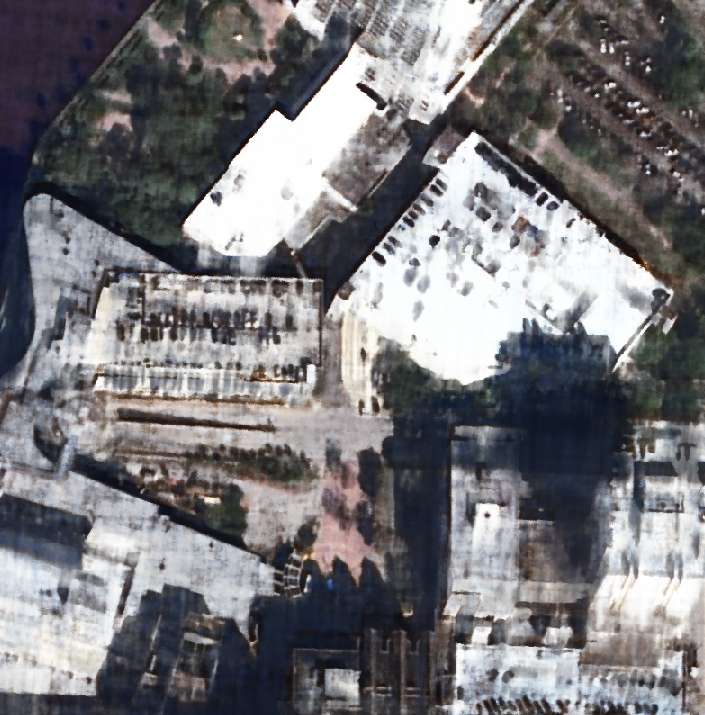}
			\end{minipage}%
		}
		\subfigure[NeRF $DFC_{3v}$]{
			\begin{minipage}[t]{0.19\linewidth}
				\centering
				\includegraphics[width=1\linewidth]{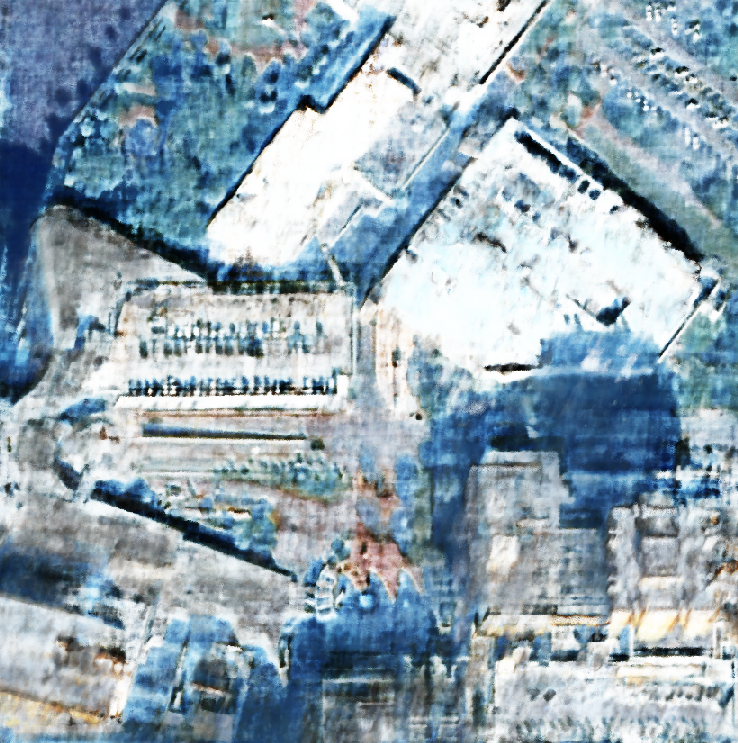}
			\end{minipage}%
		}
		\subfigure[NeRF $Dji_{2v}$]{
			\begin{minipage}[t]{0.19\linewidth}
				\centering
                    \begin{tikzpicture}
                    \node[anchor=south west,inner sep=0] (image) at (0,0) {
				\includegraphics[width=1\linewidth]{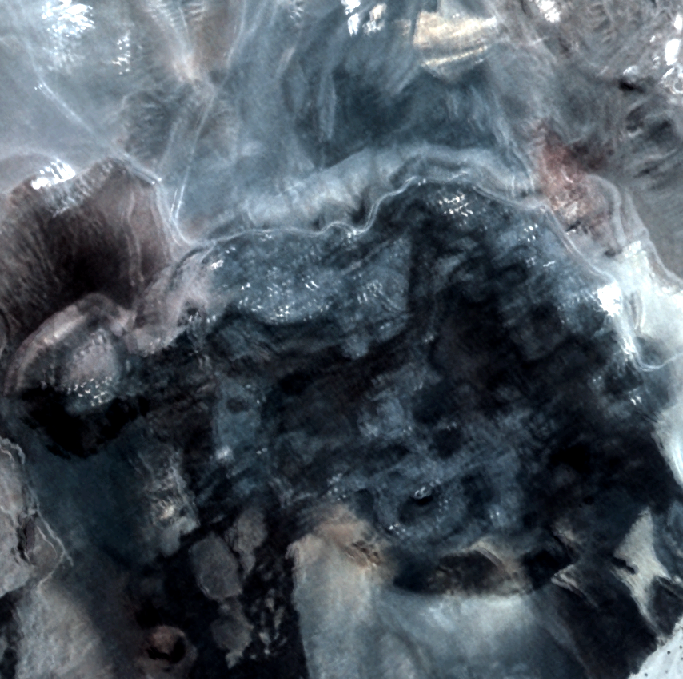}};
                    \begin{scope}[x={(image.south east)},y={(image.north west)}]
                    \draw[red,thick] (0.478,0.609) rectangle (0.622,0.748);
                    \end{scope}
                    \end{tikzpicture}
			\end{minipage}%
			\begin{minipage}[t]{0.13\linewidth}
				\centering
                    \begin{tikzpicture}
                    \node[anchor=south west,inner sep=0] (image) at (0,0) {
				\includegraphics[width=1\linewidth]{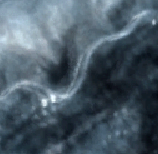}};
                    \begin{scope}[x={(image.south east)},y={(image.north west)}]
                    \draw[red,thick] (0,0) rectangle (1,1);
                    \end{scope}
                    \end{tikzpicture}
			\end{minipage}%
		}
		\subfigure[NeRF $Dji_{3v}$]{
			\begin{minipage}[t]{0.19\linewidth}
				\centering
				\includegraphics[width=1\linewidth]{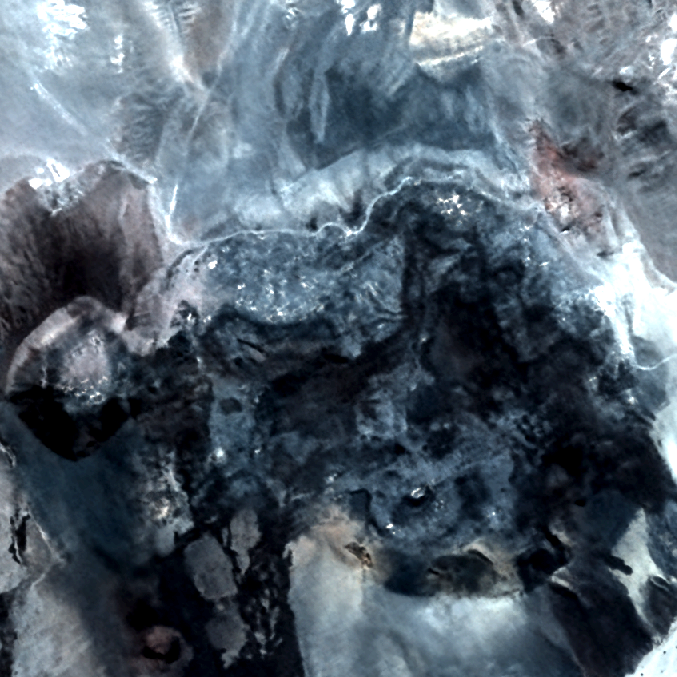}
			\end{minipage}%
		}\\

		\subfigure[SatNeRF $DFC_{2v}$]{
			\begin{minipage}[t]{0.19\linewidth}
				\centering
				\includegraphics[width=1\linewidth]{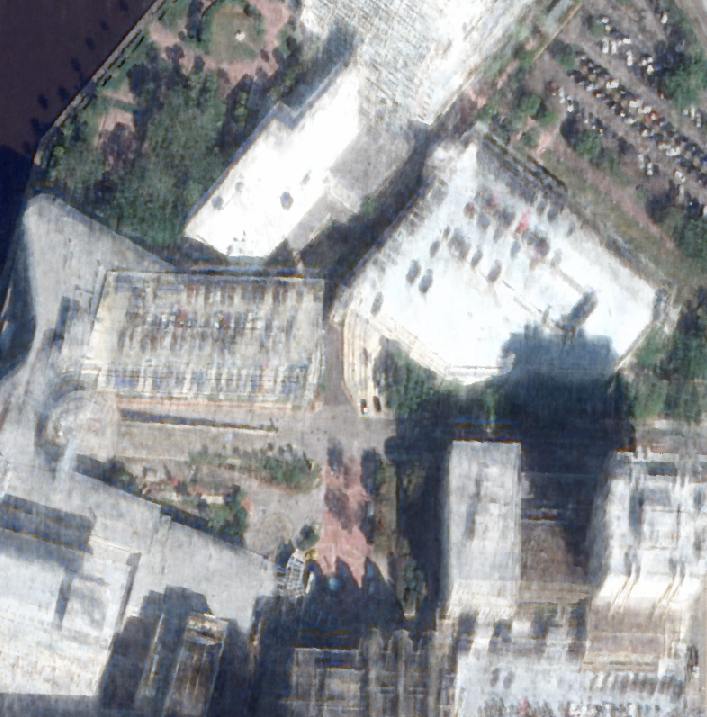}
			\end{minipage}%
		}
		\subfigure[SatNeRF $DFC_{3v}$]{
			\begin{minipage}[t]{0.19\linewidth}
				\centering
				\includegraphics[width=1\linewidth]{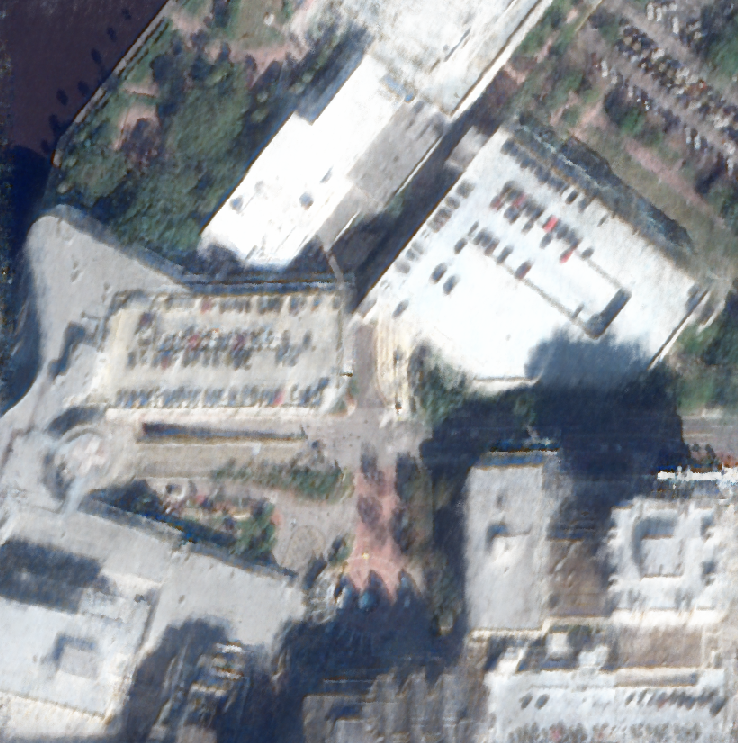}
			\end{minipage}%
		}
		\subfigure[SatNeRF $Dji_{2v}$]{
			\begin{minipage}[t]{0.19\linewidth}
				\centering
				                    \begin{tikzpicture}
                    \node[anchor=south west,inner sep=0] (image) at (0,0) {
				\includegraphics[width=1\linewidth]{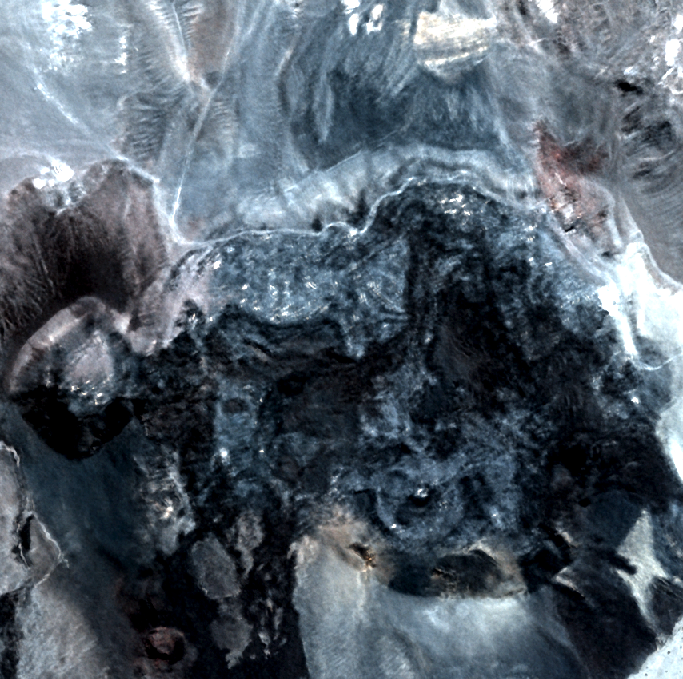}};
                    \begin{scope}[x={(image.south east)},y={(image.north west)}]
                    \draw[red,thick] (0.478,0.609) rectangle (0.622,0.748);
                    \end{scope}
                    \end{tikzpicture}
			\end{minipage}%
			\begin{minipage}[t]{0.13\linewidth}
				\centering
                    \begin{tikzpicture}
                    \node[anchor=south west,inner sep=0] (image) at (0,0) {
				\includegraphics[width=1\linewidth]{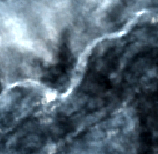}};
                    \begin{scope}[x={(image.south east)},y={(image.north west)}]
                    \draw[red,thick] (0,0) rectangle (1,1);
                    \end{scope}
                    \end{tikzpicture}
			\end{minipage}%
		}		
		\subfigure[SatNeRF $Dji_{3v}$]{
			\begin{minipage}[t]{0.19\linewidth}
				\centering
				\includegraphics[width=1\linewidth]{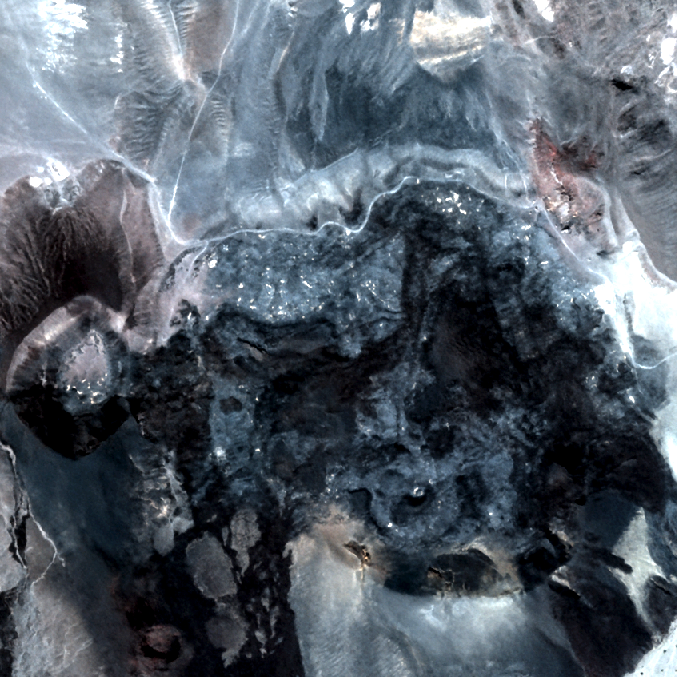}
			\end{minipage}%
		}\\

		\subfigure[\OurNeRFShort~ $DFC_{2v}$]{
			\begin{minipage}[t]{0.19\linewidth}
				\centering
				\includegraphics[width=1\linewidth]{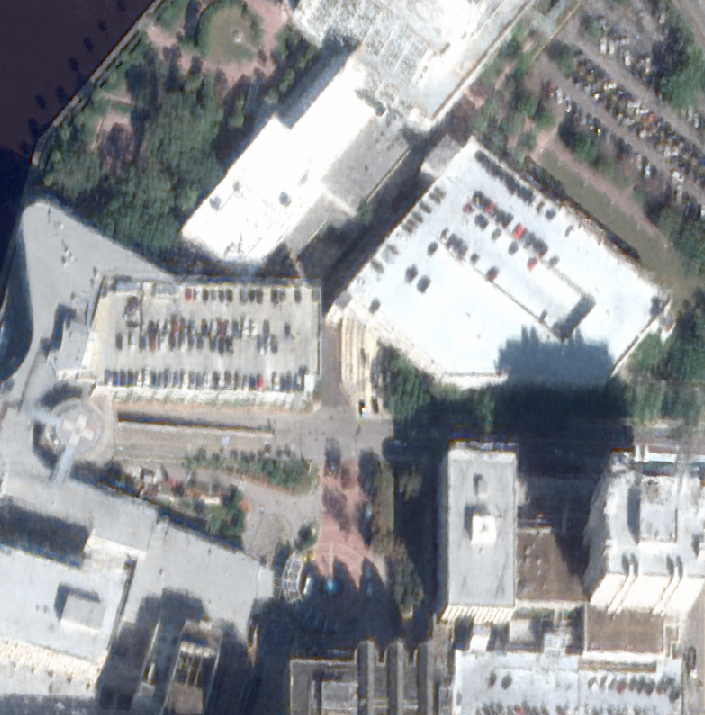}
			\end{minipage}%
		}
		\subfigure[\OurNeRFShort~ $DFC_{3v}$]{
			\begin{minipage}[t]{0.19\linewidth}
				\centering
				\includegraphics[width=1\linewidth]{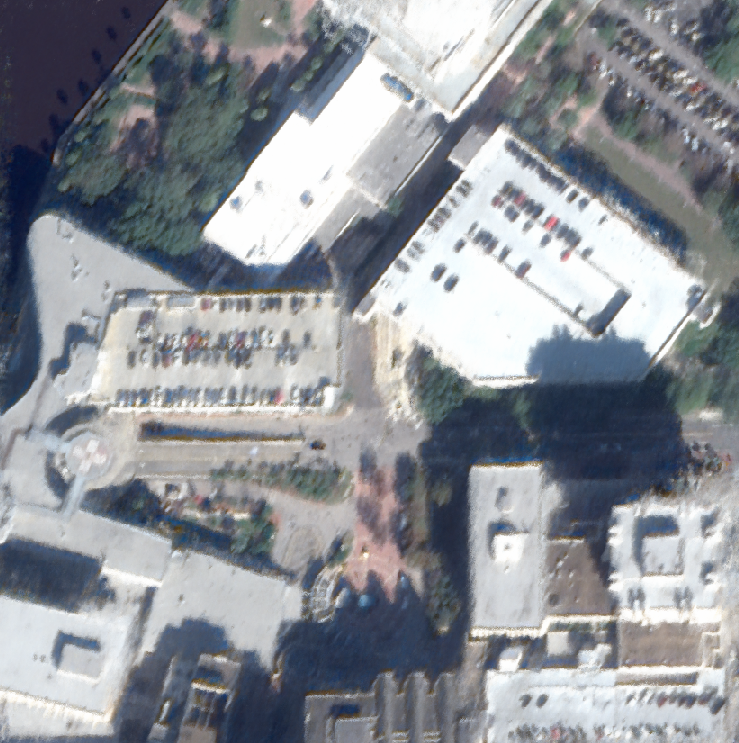}
			\end{minipage}%
		}
		\subfigure[\OurNeRFShort~ $Dji_{2v}$]{
			\begin{minipage}[t]{0.19\linewidth}
				\centering
				                    \begin{tikzpicture}
                    \node[anchor=south west,inner sep=0] (image) at (0,0) {
				\includegraphics[width=1\linewidth]{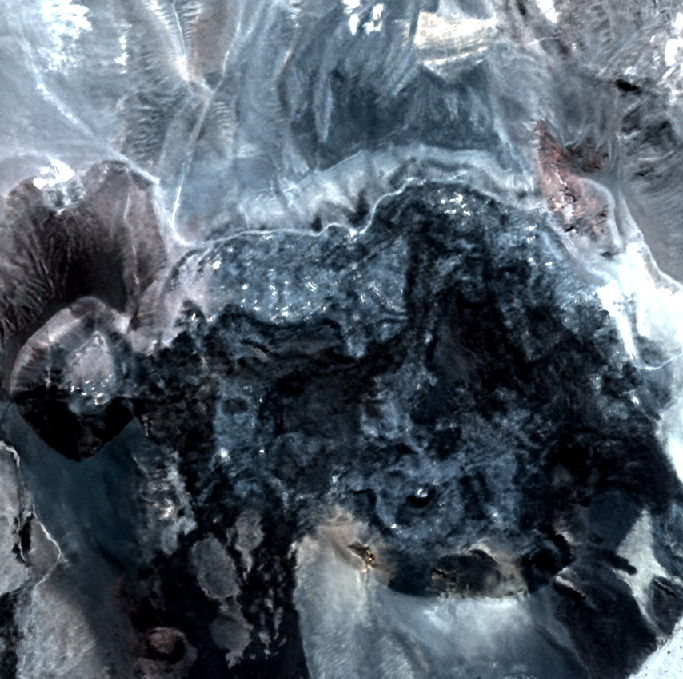}};
                    \begin{scope}[x={(image.south east)},y={(image.north west)}]
                    \draw[red,thick] (0.478,0.609) rectangle (0.622,0.748);
                    \end{scope}
                    \end{tikzpicture}
			\end{minipage}%
			\begin{minipage}[t]{0.13\linewidth}
				\centering
                    \begin{tikzpicture}
                    \node[anchor=south west,inner sep=0] (image) at (0,0) {
				\includegraphics[width=1\linewidth]{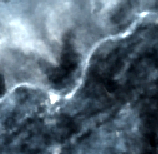}};
                    \begin{scope}[x={(image.south east)},y={(image.north west)}]
                    \draw[red,thick] (0,0) rectangle (1,1);
                    \end{scope}
                    \end{tikzpicture}
			\end{minipage}%
		}		
		\subfigure[\OurNeRFShort~ $Dji_{3v}$]{
			\begin{minipage}[t]{0.19\linewidth}
				\centering
				\includegraphics[width=1\linewidth]{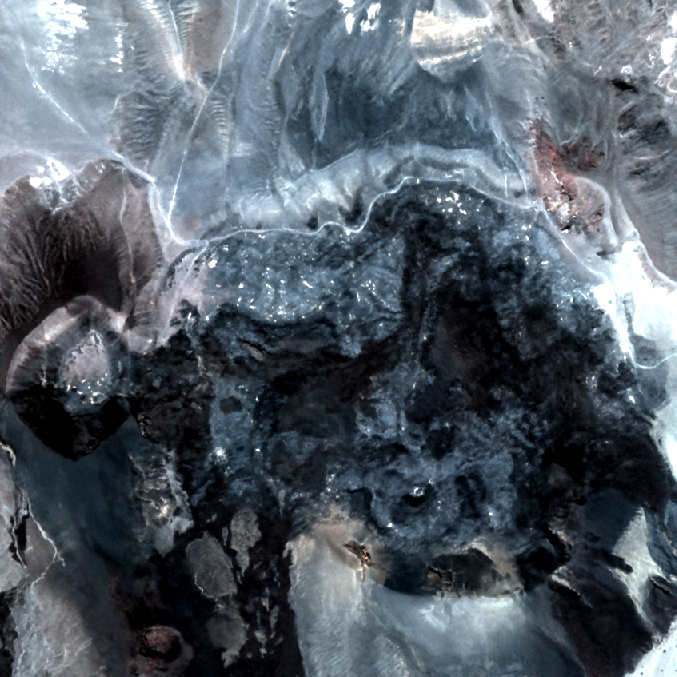}
			\end{minipage}%
		}\\

		\subfigure[GT $DFC_{2v}$]{
			\begin{minipage}[t]{0.19\linewidth}
				\centering
				\includegraphics[width=1\linewidth]{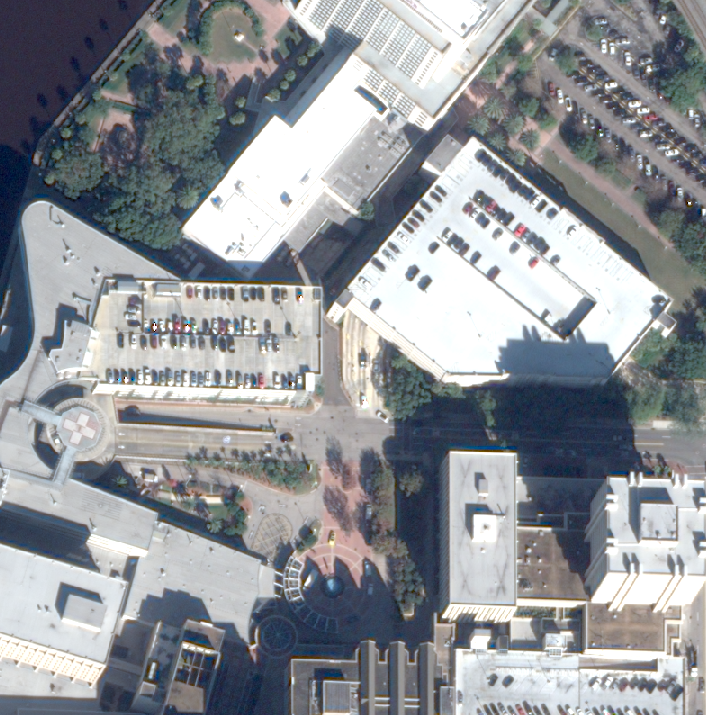}
			\end{minipage}%
		}
		\subfigure[GT $DFC_{3v}$]{
			\begin{minipage}[t]{0.19\linewidth}
				\centering
				\includegraphics[width=1\linewidth]{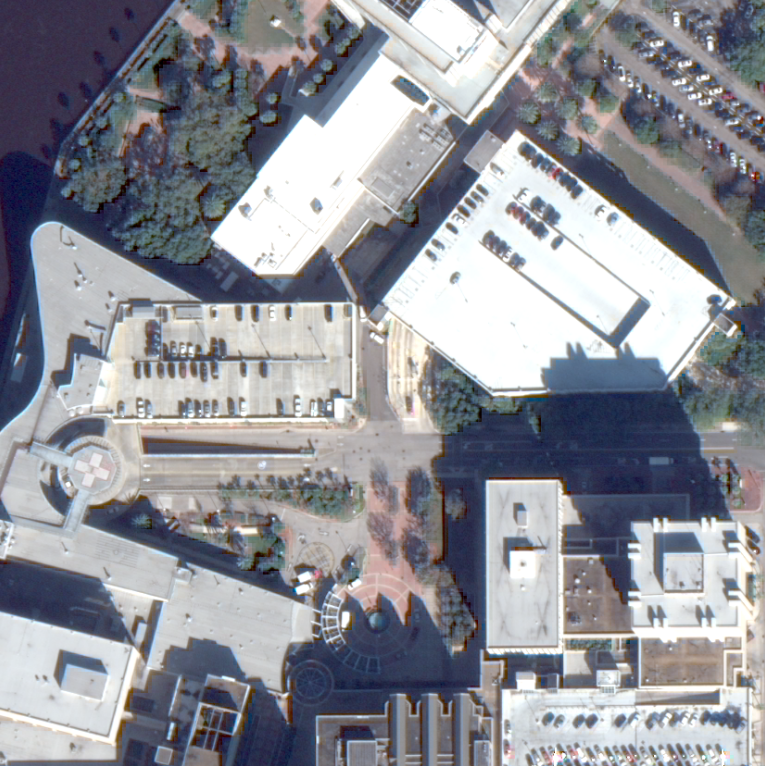}
			\end{minipage}%
		}
		\subfigure[GT $Dji_{2v}$]{
			\begin{minipage}[t]{0.19\linewidth}
				\centering
				                    \begin{tikzpicture}
                    \node[anchor=south west,inner sep=0] (image) at (0,0) {
				\includegraphics[width=1\linewidth]{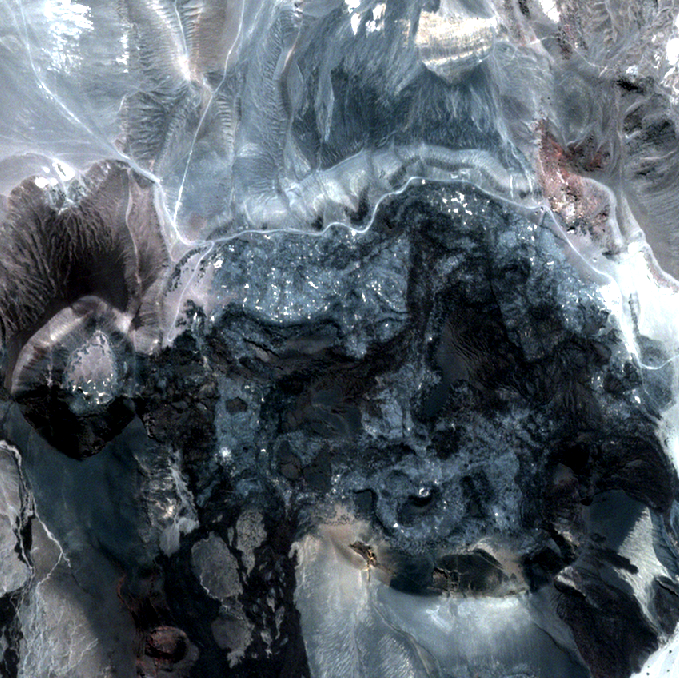}};
                    \begin{scope}[x={(image.south east)},y={(image.north west)}]
                    \draw[red,thick] (0.478,0.609) rectangle (0.622,0.748);
                    \end{scope}
                    \end{tikzpicture}
			\end{minipage}%
			\begin{minipage}[t]{0.13\linewidth}
				\centering
                    \begin{tikzpicture}
                    \node[anchor=south west,inner sep=0] (image) at (0,0) {
				\includegraphics[width=1\linewidth]{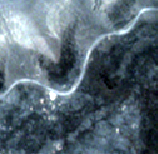}};
                    \begin{scope}[x={(image.south east)},y={(image.north west)}]
                    \draw[red,thick] (0,0) rectangle (1,1);
                    \end{scope}
                    \end{tikzpicture}
			\end{minipage}%
		}
		\subfigure[GT $Dji_{3v}$]{
			\begin{minipage}[t]{0.19\linewidth}
				\centering
				\includegraphics[width=1\linewidth]{figures/result/GT34-RGB.png}
			\end{minipage}%
		}\\

		\caption{\textbf{Novel view synthesis}. Qualitative evaluation is performed on DFC2019 (DFC) and Djibouti (Dji) datasets {using 2-views ($_{2v}$) and 3-views ($_{3v}$) for training.} \Nerf~renders blurry images, Sat-\Nerf~reduces the blur thanks to sparse depth supervision, \OurNeRFShort~renders sharpest images of all.}
		\label{RGBComp}
	\end{center}
\end{figure*}

\begin{figure*}[htbp]
	\begin{center}

		\subfigure[NeRF $DFC_{2v}$]{
			\begin{minipage}[t]{0.18\linewidth}
				\centering
				\includegraphics[width=1\linewidth]{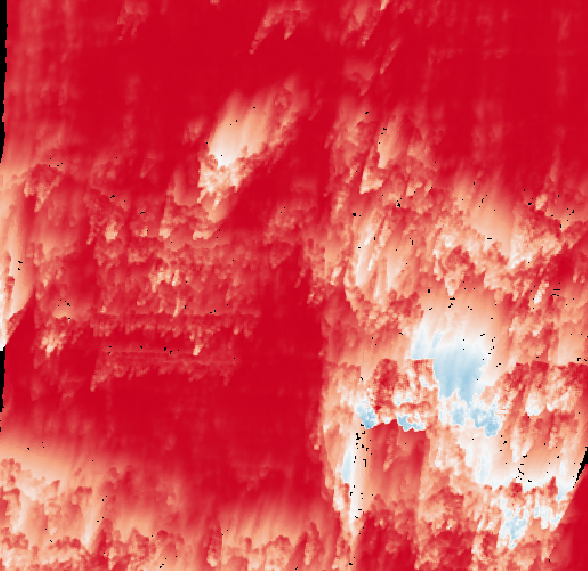}
			\end{minipage}%
		}
		\subfigure[NeRF $DFC_{3v}$]{
			\begin{minipage}[t]{0.18\linewidth}
				\centering
				\includegraphics[width=1\linewidth]{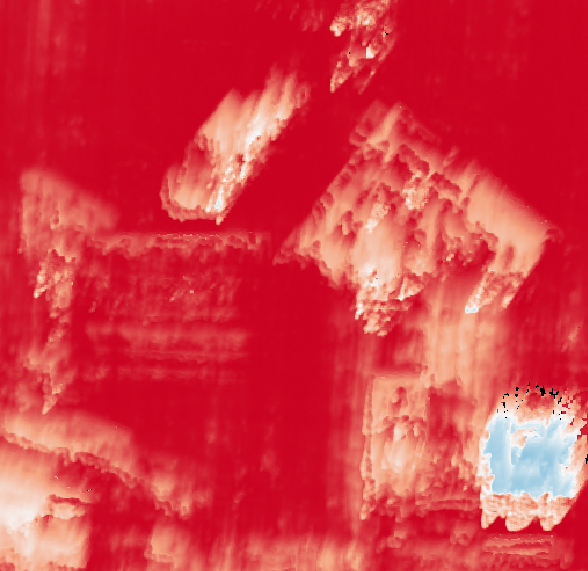}
			\end{minipage}%
		}
		\subfigure[NeRF $Dji_{2v}$]{
			\begin{minipage}[t]{0.18\linewidth}
				\centering
                        \begin{tikzpicture}
                    \node[anchor=south west,inner sep=0] (image) at (0,0) {
				\includegraphics[width=1\linewidth]{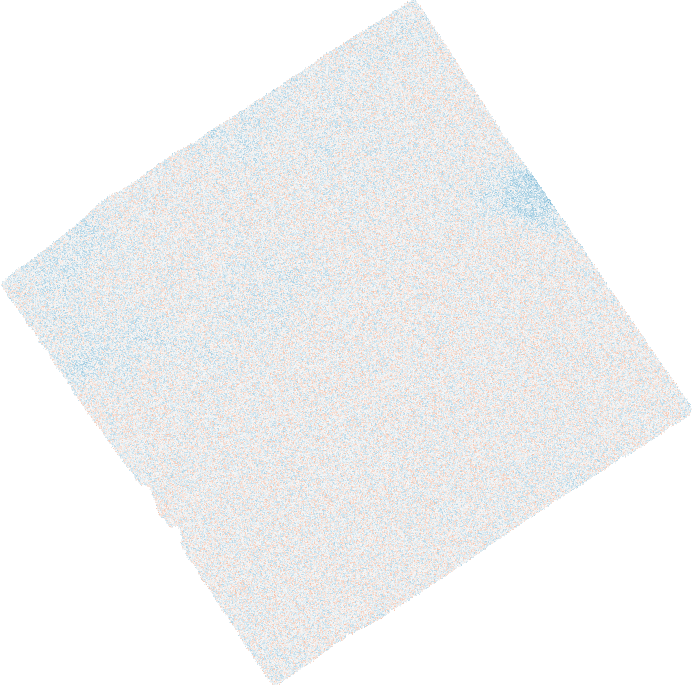}};
                    \begin{scope}[x={(image.south east)},y={(image.north west)}]
                    \draw[red,thick] (0.838,0.464) rectangle (0.897,0.522);
                    \end{scope}
                    \end{tikzpicture}                          
			\end{minipage}%
			\begin{minipage}[t]{0.13\linewidth}
				\centering
                    \begin{tikzpicture}
                    \node[anchor=south west,inner sep=0] (image) at (0,0) {
				\includegraphics[width=1\linewidth]{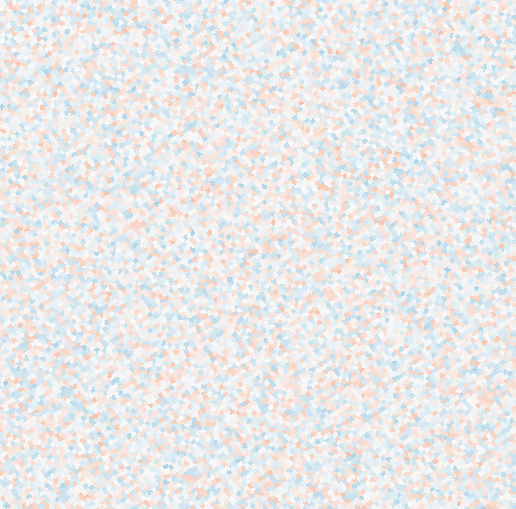}};
                    \begin{scope}[x={(image.south east)},y={(image.north west)}]
                    \draw[red,thick] (0,0) rectangle (1,1);
                    \end{scope}
                    \end{tikzpicture}
			\end{minipage}%
		}      
		\subfigure[NeRF $Dji_{3v}$]{
			\begin{minipage}[t]{0.18\linewidth}
				\centering
				\includegraphics[width=1\linewidth]{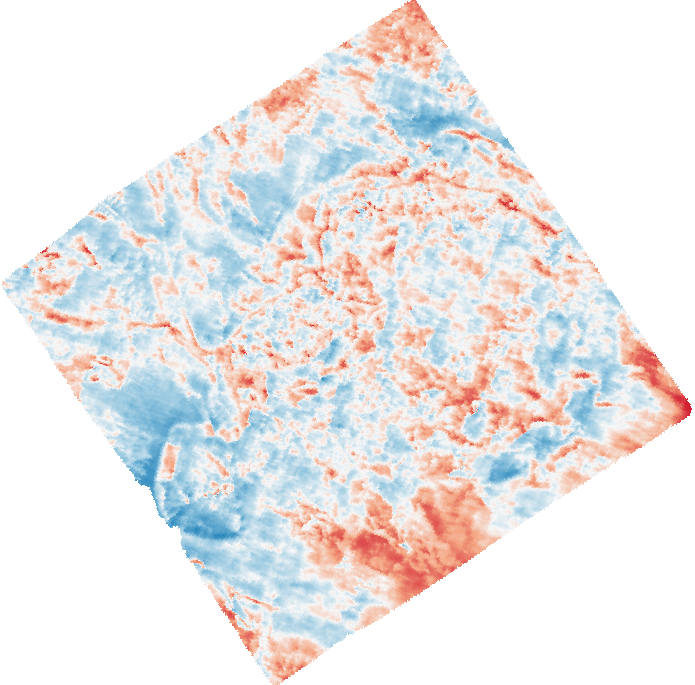}
			\end{minipage}%
		}\\

		\subfigure[SatNeRF $DFC_{2v}$]{
			\begin{minipage}[t]{0.18\linewidth}
				\centering
				\includegraphics[width=1\linewidth]{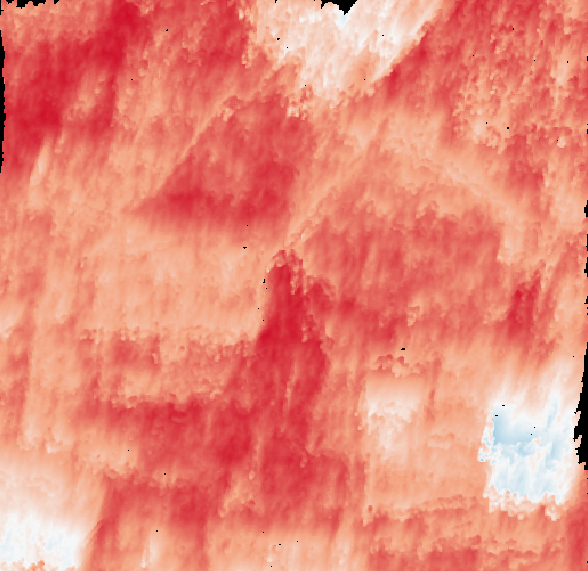}
			\end{minipage}%
		}
		\subfigure[SatNeRF $DFC_{3v}$]{
			\begin{minipage}[t]{0.18\linewidth}
				\centering
				\includegraphics[width=1\linewidth]{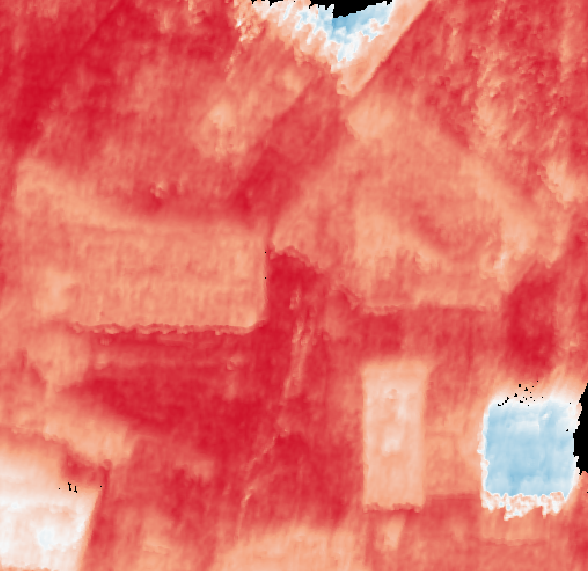}
			\end{minipage}%
		}
		\subfigure[SatNeRF $Dji_{2v}$]{
			\begin{minipage}[t]{0.18\linewidth}
				\centering
                        \begin{tikzpicture}
                    \node[anchor=south west,inner sep=0] (image) at (0,0) {
				\includegraphics[width=1\linewidth]{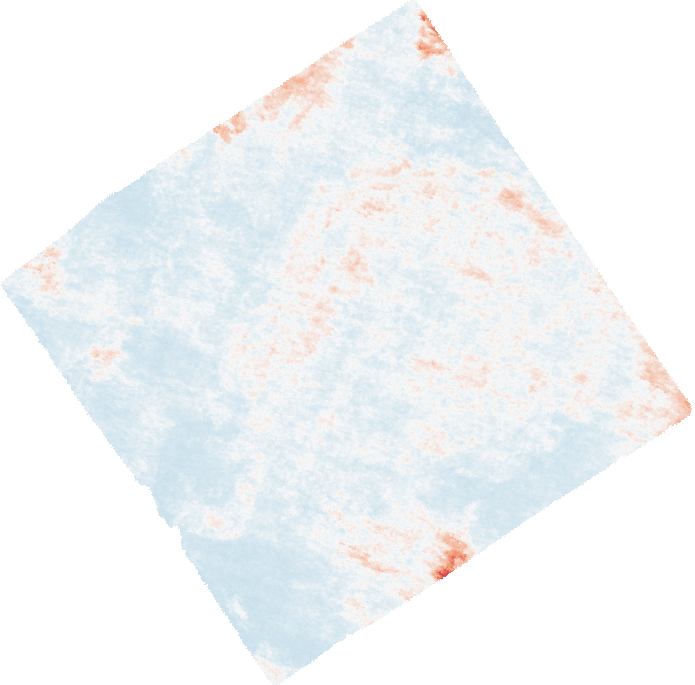}};
                    \begin{scope}[x={(image.south east)},y={(image.north west)}]
                    \draw[red,thick] (0.838,0.464) rectangle (0.897,0.522);
                    \end{scope}
                    \end{tikzpicture}                      
			\end{minipage}%
			\begin{minipage}[t]{0.13\linewidth}
				\centering
                    \begin{tikzpicture}
                    \node[anchor=south west,inner sep=0] (image) at (0,0) {
				\includegraphics[width=1\linewidth]{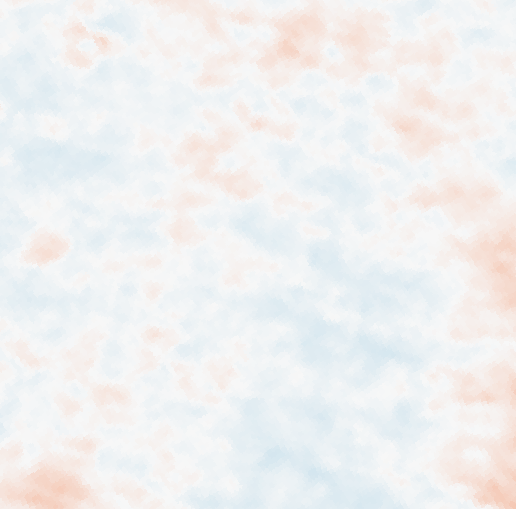}};
                    \begin{scope}[x={(image.south east)},y={(image.north west)}]
                    \draw[red,thick] (0,0) rectangle (1,1);
                    \end{scope}
                    \end{tikzpicture}
			\end{minipage}%
		}      
		\subfigure[SatNeRF $Dji_{3v}$]{
			\begin{minipage}[t]{0.18\linewidth}
				\centering
				\includegraphics[width=1\linewidth]{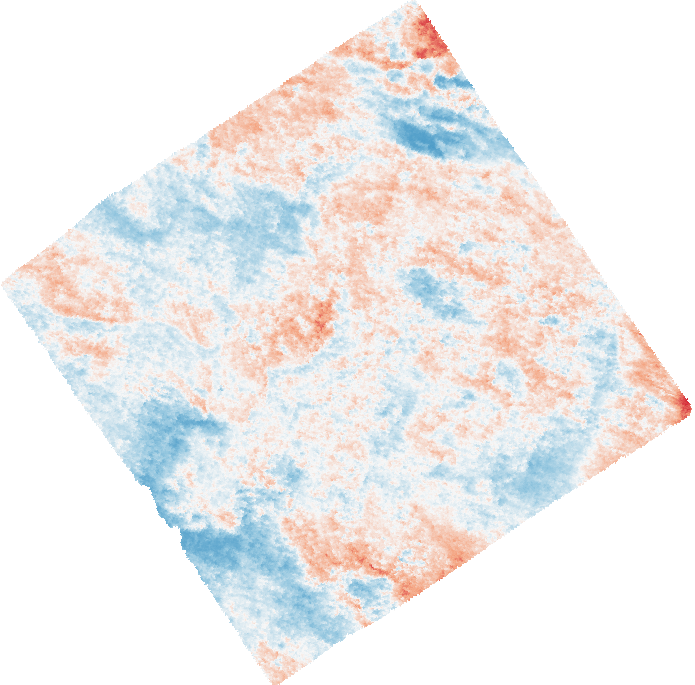}
			\end{minipage}%
		}\\

		\subfigure[$SGM_{scl4}$ $DFC_{2v}$]{
			\begin{minipage}[t]{0.18\linewidth}
				\centering
				\includegraphics[width=1\linewidth]{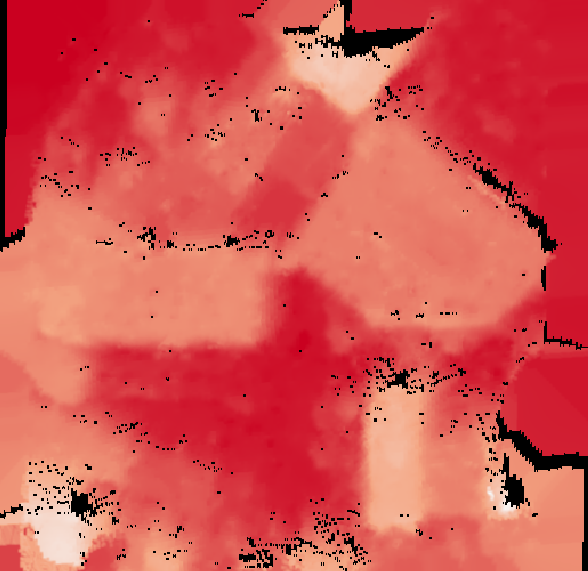}
			\end{minipage}%
		}
		\subfigure[$SGM_{scl4}$ $DFC_{3v}$]{
			\begin{minipage}[t]{0.18\linewidth}
				\centering
				\includegraphics[width=1\linewidth]{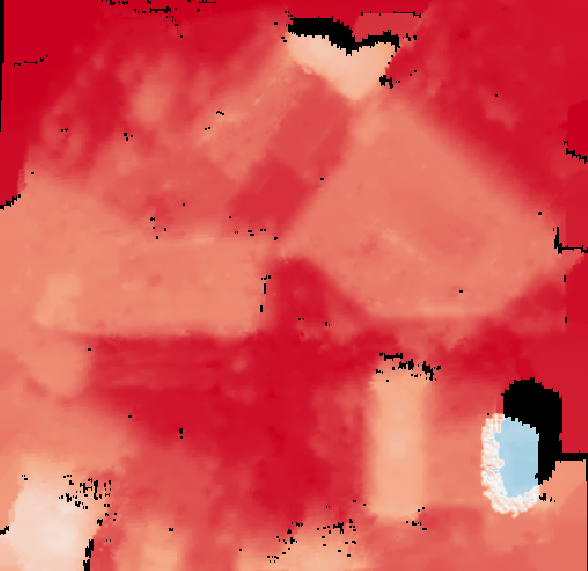}
			\end{minipage}%
		}
		\subfigure[$SGM_{scl4}$ $Dji_{2v}$]{
			\begin{minipage}[t]{0.18\linewidth}
				\centering
                        \begin{tikzpicture}
                    \node[anchor=south west,inner sep=0] (image) at (0,0) {
				\includegraphics[width=1\linewidth]{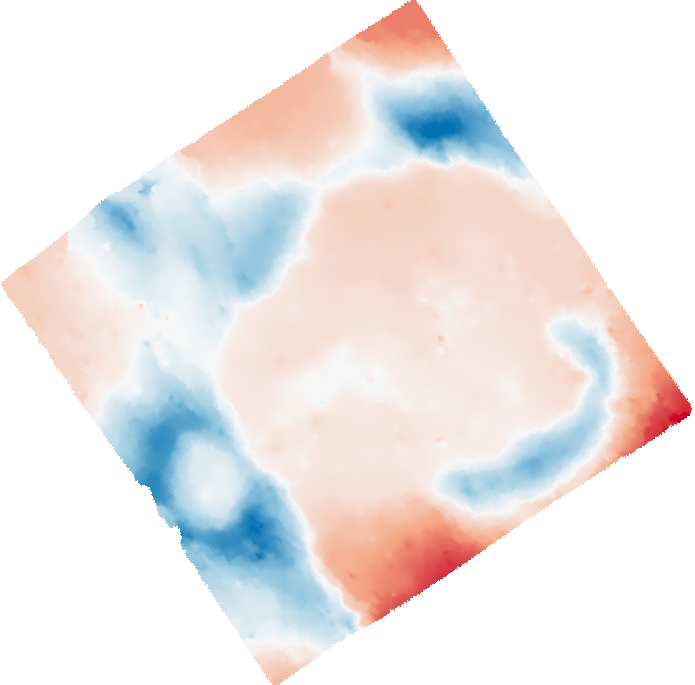}};
                    \begin{scope}[x={(image.south east)},y={(image.north west)}]
                    \draw[red,thick] (0.838,0.464) rectangle (0.897,0.522);
                    \end{scope}
                    \end{tikzpicture}                  
			\end{minipage}%
			\begin{minipage}[t]{0.13\linewidth}
				\centering
                    \begin{tikzpicture}
                    \node[anchor=south west,inner sep=0] (image) at (0,0) {
				\includegraphics[width=1\linewidth]{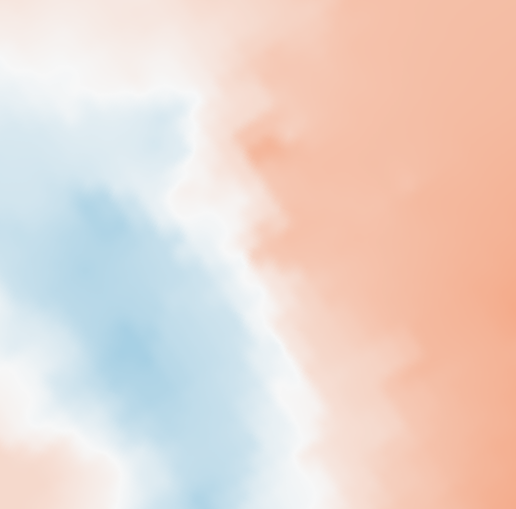}};
                    \begin{scope}[x={(image.south east)},y={(image.north west)}]
                    \draw[red,thick] (0,0) rectangle (1,1);
                    \end{scope}
                    \end{tikzpicture}
			\end{minipage}%
		}    
		\subfigure[$SGM_{scl4}$ $Dji_{3v}$]{
			\begin{minipage}[t]{0.18\linewidth}
				\centering
				\includegraphics[width=1\linewidth]{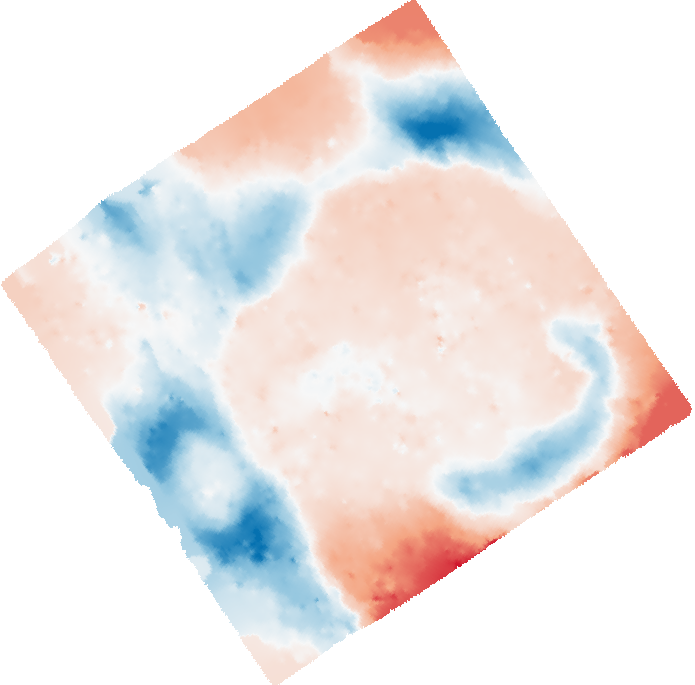}
			\end{minipage}%
		}\\
  
		\subfigure[\OurNeRFShort~ $DFC_{2v}$]{
			\begin{minipage}[t]{0.18\linewidth}
				\centering
                        \begin{tikzpicture}
                    \node[anchor=south west,inner sep=0] (image) at (0,0) {
				\includegraphics[width=1\linewidth]{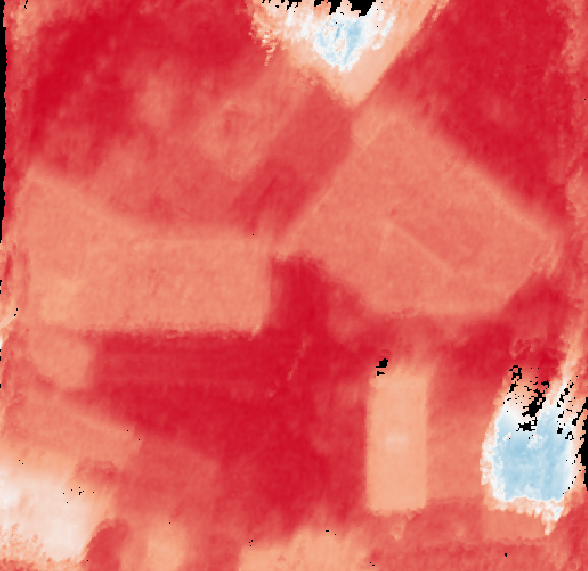}};
                    \begin{scope}[x={(image.south east)},y={(image.north west)}]
                    \draw[yellow,thick] (0.594,0.083) rectangle (0.654,0.267);
                    \draw[yellow,thick] (0.807,0.090) rectangle (0.964,0.275);
                    \end{scope}
                    \end{tikzpicture}
			\end{minipage}%
		}
		\subfigure[\OurNeRFShort~ $DFC_{3v}$]{
			\begin{minipage}[t]{0.18\linewidth}
				\centering
                        \begin{tikzpicture}
                    \node[anchor=south west,inner sep=0] (image) at (0,0) {
				\includegraphics[width=1\linewidth]{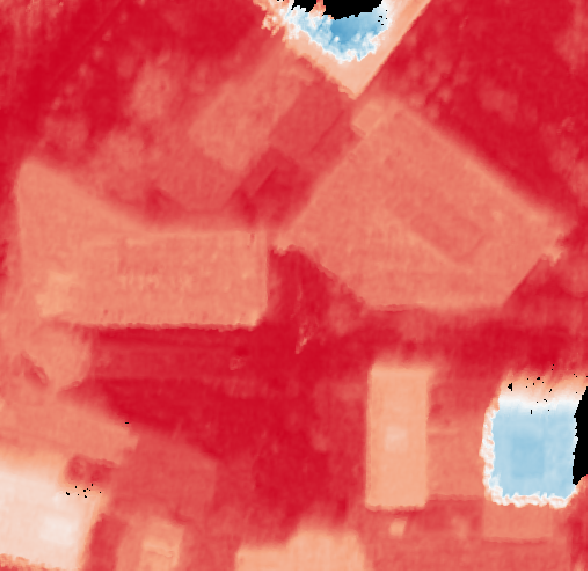}};
                    \begin{scope}[x={(image.south east)},y={(image.north west)}]
                    \draw[yellow,thick] (0.594,0.083) rectangle (0.654,0.267);
                    \draw[yellow,thick] (0.807,0.090) rectangle (0.964,0.275);
                \draw[rotate around={-42:(0.736,0.951)}, green, thick] (0.736,0.951) rectangle (0.856,0.802);
                \draw[rotate around={-42:(0.20,1)}, green, thick] (0.20,1) rectangle (0.270,0.724);
                    \end{scope}
                    \end{tikzpicture}    
			\end{minipage}%
		}
		\subfigure[\OurNeRFShort~ $Dji_{2v}$]{
			\begin{minipage}[t]{0.18\linewidth}
				\centering
                        \begin{tikzpicture}
                    \node[anchor=south west,inner sep=0] (image) at (0,0) {
                \includegraphics[width=1\linewidth]{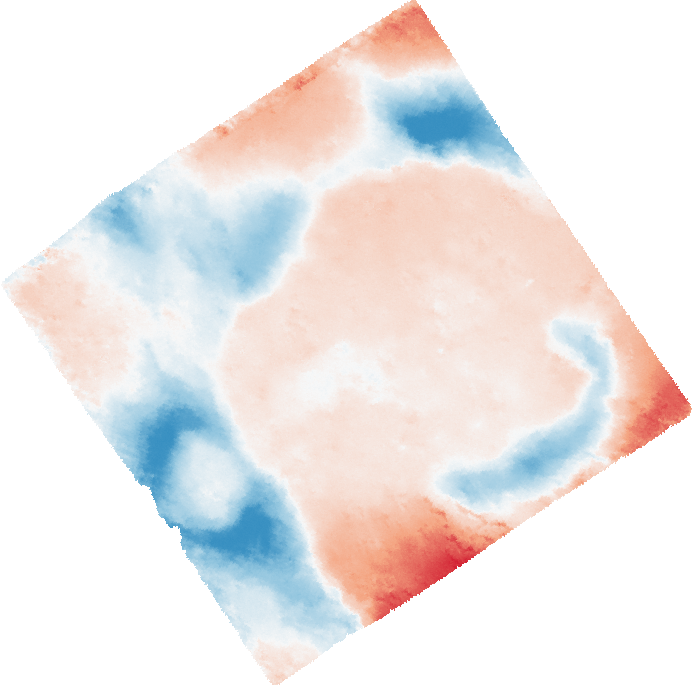}};
                    \begin{scope}[x={(image.south east)},y={(image.north west)}]
                    \draw[red,thick] (0.838,0.464) rectangle (0.897,0.522);
                    \end{scope}
                    \end{tikzpicture}        
			\end{minipage}%
			\begin{minipage}[t]{0.13\linewidth}
				\centering
                    \begin{tikzpicture}
                    \node[anchor=south west,inner sep=0] (image) at (0,0) {
				\includegraphics[width=1\linewidth]{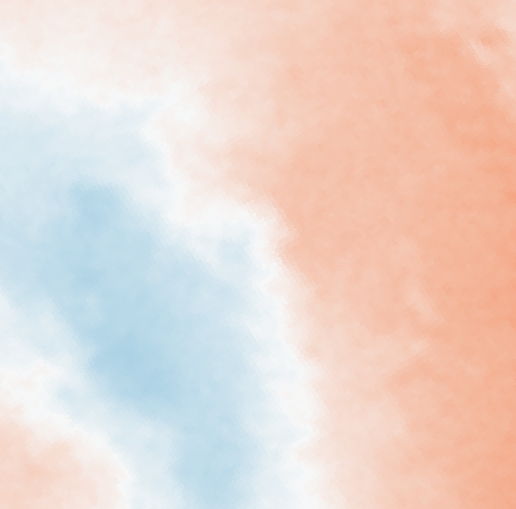}};
                    \begin{scope}[x={(image.south east)},y={(image.north west)}]
                    \draw[red,thick] (0,0) rectangle (1,1);
                    \end{scope}
                    \end{tikzpicture}
			\end{minipage}%
		}    
		\subfigure[\OurNeRFShort~ $Dji_{3v}$]{
			\begin{minipage}[t]{0.18\linewidth}
				\centering
				\includegraphics[width=1\linewidth]{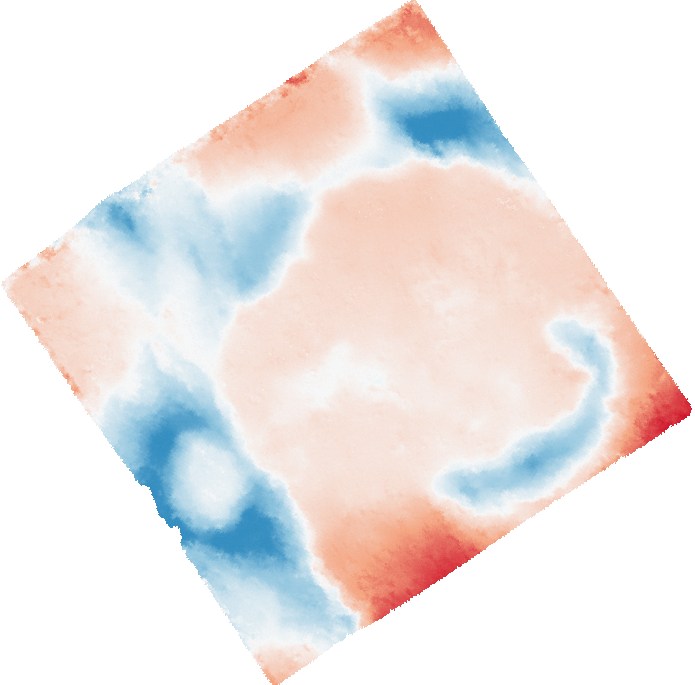}
			\end{minipage}%
		}\\

  		\subfigure[$SGM_{scl1}$ $DFC_{2v}$]{
			\begin{minipage}[t]{0.18\linewidth}
				\centering
                        \begin{tikzpicture}
                    \node[anchor=south west,inner sep=0] (image) at (0,0) {
				\includegraphics[width=1\linewidth]{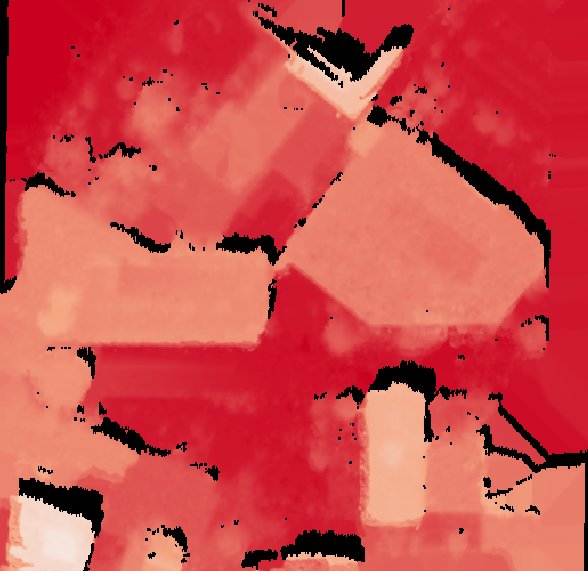}};
                    \begin{scope}[x={(image.south east)},y={(image.north west)}]
                    \draw[yellow,thick] (0.594,0.083) rectangle (0.654,0.267);
                    \draw[yellow,thick] (0.807,0.090) rectangle (0.964,0.275);
                    \end{scope}
                    \end{tikzpicture}
			\end{minipage}%
		}
		\subfigure[$SGM_{scl1}$ $DFC_{3v}$]{
			\begin{minipage}[t]{0.18\linewidth}
				\centering
                        \begin{tikzpicture}
                    \node[anchor=south west,inner sep=0] (image) at (0,0) {
				\includegraphics[width=1\linewidth]{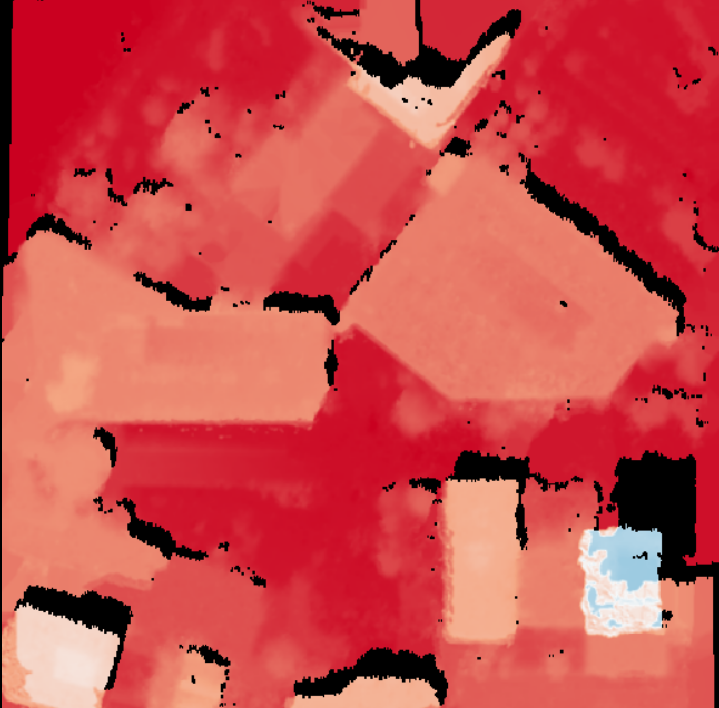}};
                    \begin{scope}[x={(image.south east)},y={(image.north west)}]
                    \draw[yellow,thick] (0.594,0.083) rectangle (0.654,0.267);
                    \draw[yellow,thick] (0.807,0.090) rectangle (0.964,0.275);
                \draw[rotate around={-42:(0.736,0.951)}, green, thick] (0.736,0.951) rectangle (0.856,0.802);
                \draw[rotate around={-42:(0.20,1)}, green, thick] (0.20,1) rectangle (0.270,0.724);
                    \end{scope}
                    \end{tikzpicture}    
			\end{minipage}%
		}
		\subfigure[$SGM_{scl1}$ $Dji_{2v}$]{
			\begin{minipage}[t]{0.18\linewidth}
				\centering
                        \begin{tikzpicture}
                    \node[anchor=south west,inner sep=0] (image) at (0,0) {
				\includegraphics[width=1\linewidth]{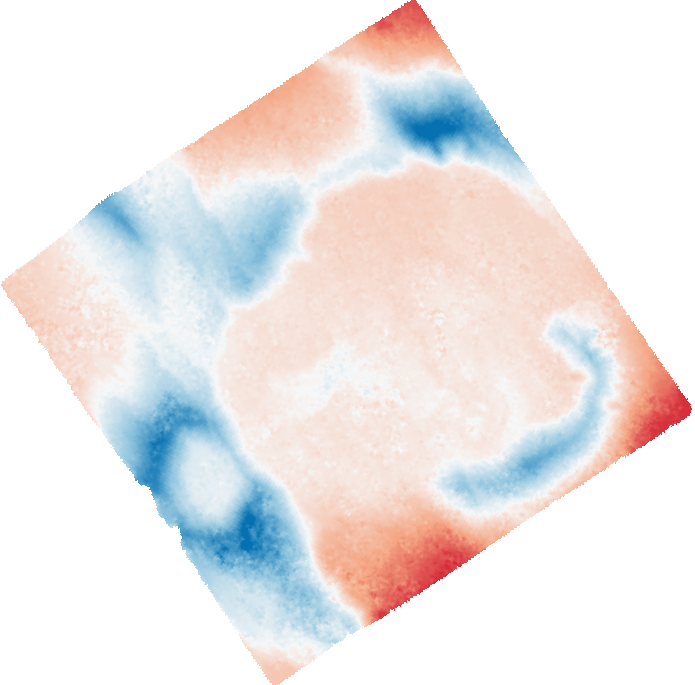}};
                    \begin{scope}[x={(image.south east)},y={(image.north west)}]
                    \draw[red,thick] (0.838,0.464) rectangle (0.897,0.522);
                    \end{scope}
                    \end{tikzpicture}                   
			\end{minipage}%
			\begin{minipage}[t]{0.13\linewidth}
				\centering
                    \begin{tikzpicture}
                    \node[anchor=south west,inner sep=0] (image) at (0,0) {
				\includegraphics[width=1\linewidth]{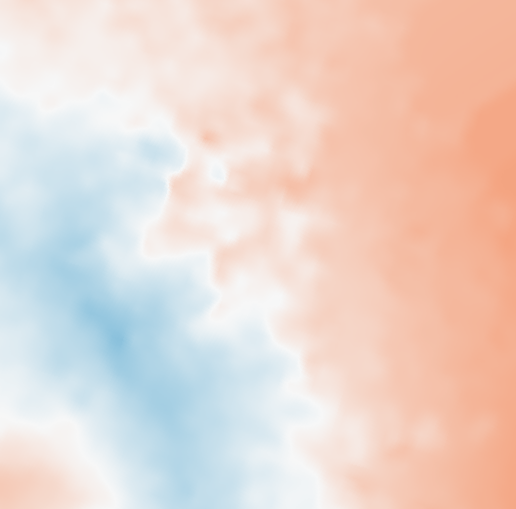}};
                    \begin{scope}[x={(image.south east)},y={(image.north west)}]
                    \draw[red,thick] (0,0) rectangle (1,1);
                    \end{scope}
                    \end{tikzpicture}
			\end{minipage}%
		}  
		\subfigure[$SGM_{scl1}$ $Dji_{3v}$]{
			\begin{minipage}[t]{0.18\linewidth}
				\centering
				\includegraphics[width=1\linewidth]{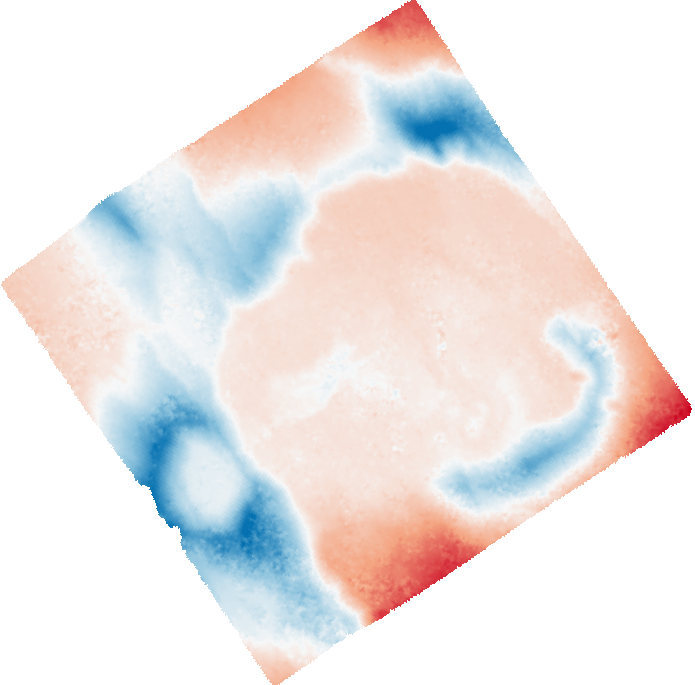}
			\end{minipage}%
		}\\

        \subfigure[GT DSM $DFC_{2v}$ ]{
            \begin{minipage}[t]{0.18\linewidth}
                \centering
                \includegraphics[width=1\linewidth]{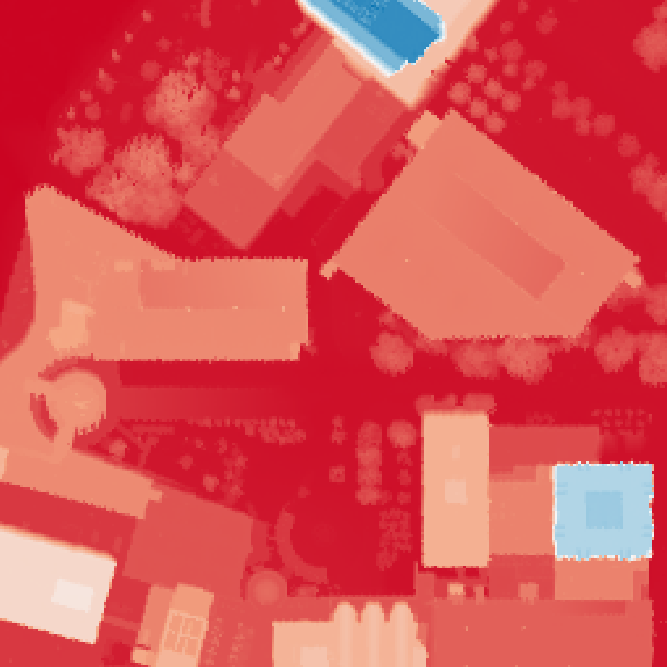}
            \end{minipage}%
        }
        \subfigure[GT DSM $DFC_{3v}$ ]{
            \begin{minipage}[t]{0.18\linewidth}
                \centering
                \includegraphics[width=1\linewidth]{figures/result/GT12-DSM.png}
            \end{minipage}%
        }
        \subfigure[GT DSM $Dji_{2v}$]{
            \begin{minipage}[t]{0.18\linewidth}
                \centering
                        \begin{tikzpicture}
                    \node[anchor=south west,inner sep=0] (image) at (0,0) {
				\includegraphics[width=1\linewidth]{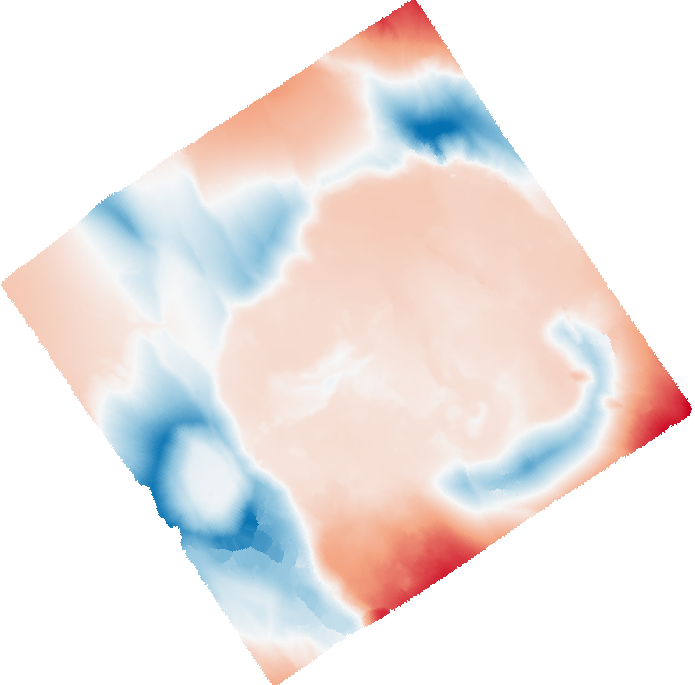}};
                    \begin{scope}[x={(image.south east)},y={(image.north west)}]
                    \draw[red,thick] (0.838,0.464) rectangle (0.897,0.522);
                    \end{scope}
                    \end{tikzpicture}                    
            \end{minipage}%
			\begin{minipage}[t]{0.13\linewidth}
				\centering
                    \begin{tikzpicture}
                    \node[anchor=south west,inner sep=0] (image) at (0,0) {
				\includegraphics[width=1\linewidth]{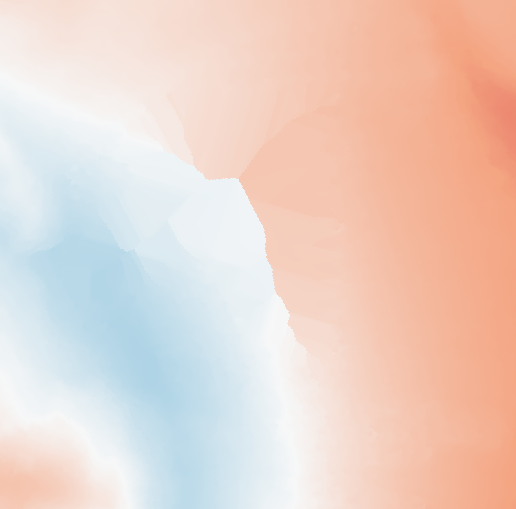}};
                    \begin{scope}[x={(image.south east)},y={(image.north west)}]
                    \draw[red,thick] (0,0) rectangle (1,1);
                    \end{scope}
                    \end{tikzpicture}
			\end{minipage}%
		}
        \subfigure[GT DSM $Dji_{3v}$]{
            \begin{minipage}[t]{0.18\linewidth}
                \centering
                \includegraphics[width=1\linewidth]{figures/result/GT34-DSM.png}
            \end{minipage}%
        }\\
        
			\begin{minipage}[t]{0.38\linewidth}
				\centering
				\includegraphics[width=1\linewidth]{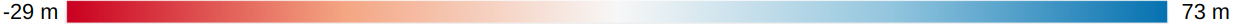}
			\end{minipage}%
			\begin{minipage}[t]{0.5\linewidth}
				\centering
				\includegraphics[width=1\linewidth]{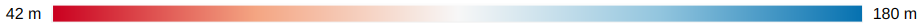}
			\end{minipage}%

		\caption{\textbf{Altitude extraction.} \OurNeRFShort~outruns all tested \Nerf~variants, and reconstructs 3D geometry comparably to SGM$_{scl1}$. In urban DFC2019 dataset, \OurNeRFShort~is better at reconstructing vegetation (\textcolor{green}{\textbf{$\square$}}) and at handling building outlines near occlusions (\textcolor{yellow}{\textbf{$\square$}}) but the surface is generally less smooth than that of SGM$_{scl1}$. In rural Djibouti dataset, notice the more detailed and coherent reconstruction of \OurNeRFShort~in (o) compared to SGM$_{scl1}$ result in (s).}
        \label{DSMComp}  
	\end{center}
\end{figure*}

\begin{figure*}[htbp]
	\begin{center}

		\subfigure[\OurNeRFShort~ $DFC_{2v}$]{
			\begin{minipage}[t]{0.18\linewidth}
				\centering
				\includegraphics[width=1\linewidth]{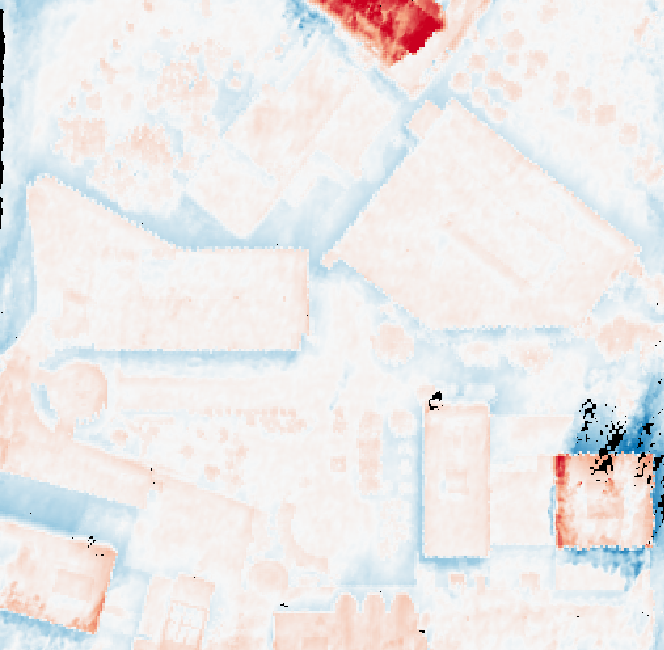}
			\end{minipage}%
		}
		\subfigure[\OurNeRFShort~ $DFC_{3v}$]{
			\begin{minipage}[t]{0.18\linewidth}
				\centering
				\includegraphics[width=1\linewidth]{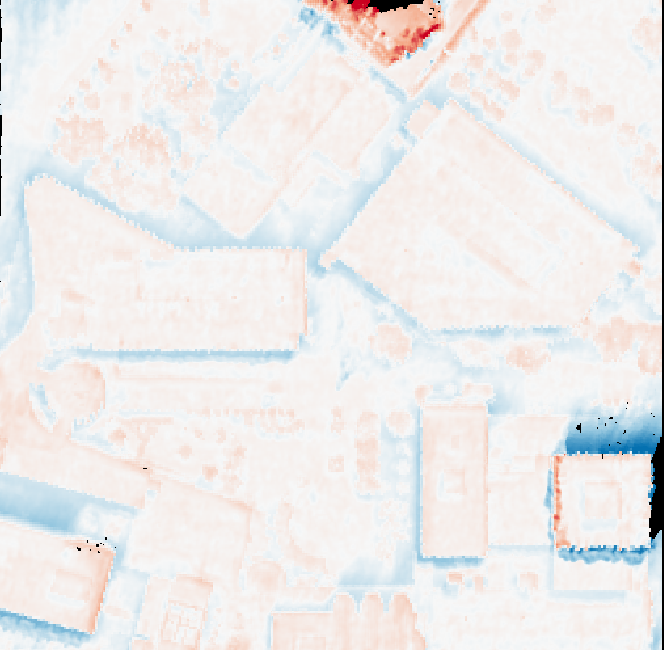}
			\end{minipage}%
		}
		\subfigure[\OurNeRFShort~ $Dji_{2v}$]{
			\begin{minipage}[t]{0.18\linewidth}
				\centering
				\includegraphics[width=1\linewidth]{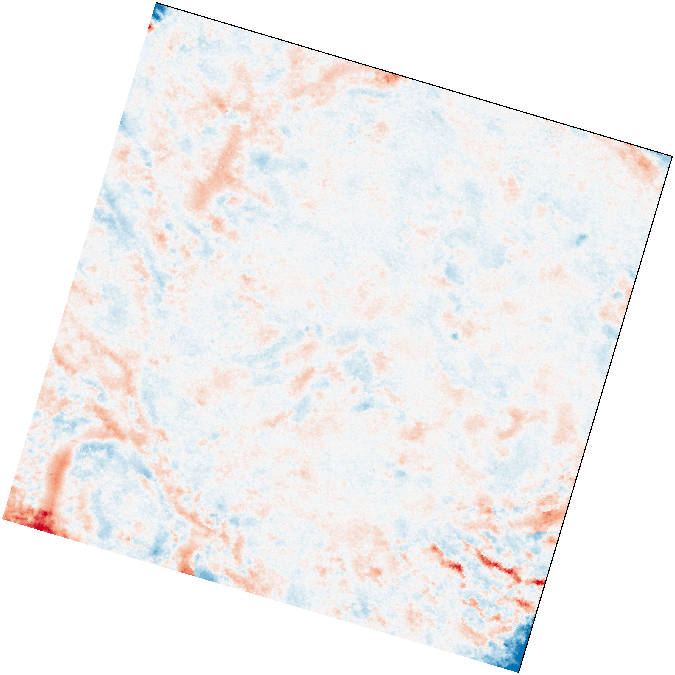}
			\end{minipage}%
		}
		\subfigure[\OurNeRFShort~ $Dji_{3v}$]{
			\begin{minipage}[t]{0.18\linewidth}
				\centering
				\includegraphics[width=1\linewidth]{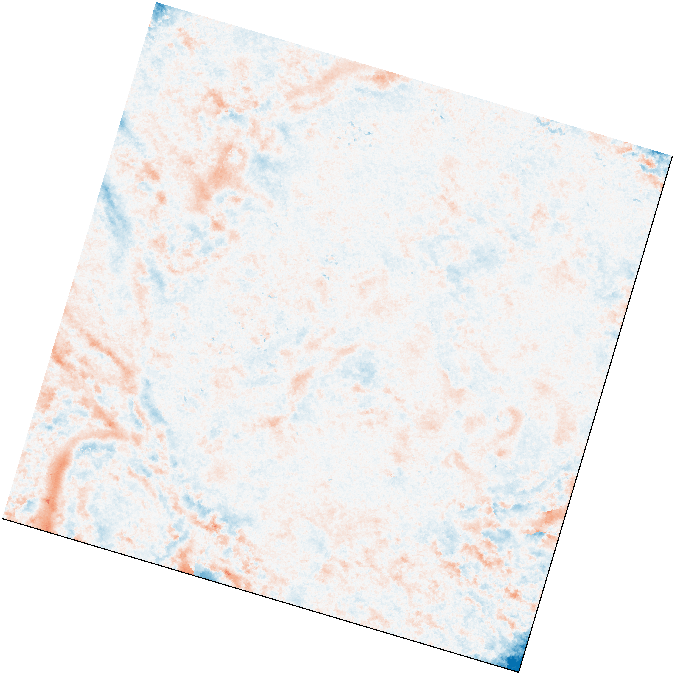}
			\end{minipage}%
		}\\

  		\subfigure[$SGM_{scl1}$ $DFC_{2v}$]{
			\begin{minipage}[t]{0.18\linewidth}
				\centering
    \includegraphics[width=1\linewidth]{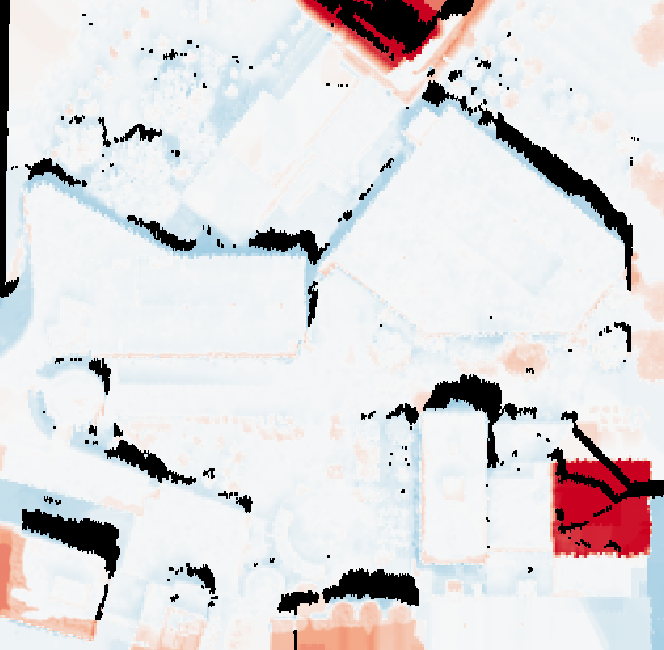}
			\end{minipage}%
		}
		\subfigure[$SGM_{scl1}$ $DFC_{3v}$]{
			\begin{minipage}[t]{0.18\linewidth}
				\centering
    \includegraphics[width=1\linewidth]{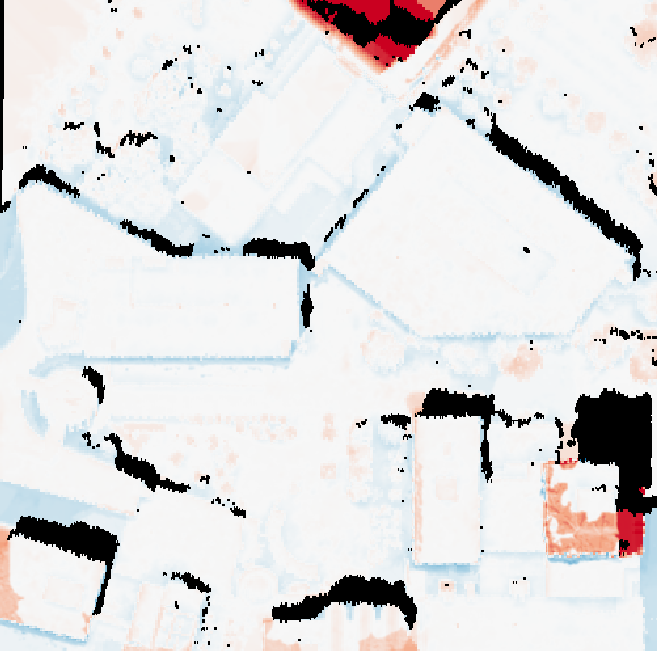}
			\end{minipage}%
		}
		\subfigure[$SGM_{scl1}$ $Dji_{2v}$]{
			\begin{minipage}[t]{0.18\linewidth}
				\centering
				\includegraphics[width=1\linewidth]{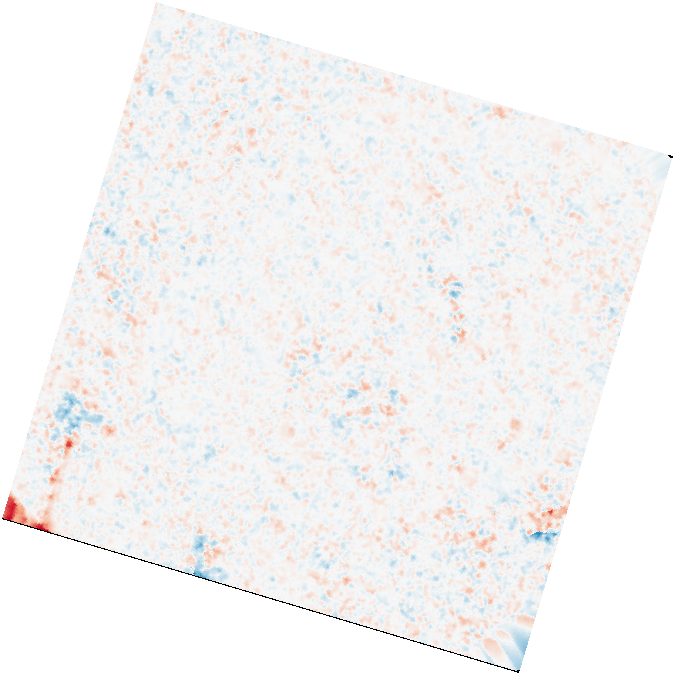}
			\end{minipage}%
		}
		\subfigure[$SGM_{scl1}$ $Dji_{3v}$]{
			\begin{minipage}[t]{0.18\linewidth}
				\centering
				\includegraphics[width=1\linewidth]{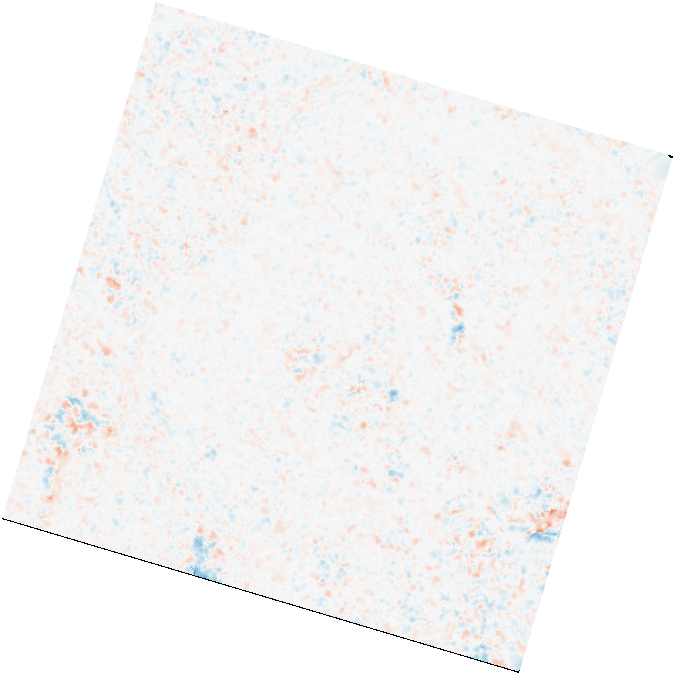}
			\end{minipage}%
		}\\
		
			\begin{minipage}[t]{0.35\linewidth}
				\centering
				\includegraphics[width=1\linewidth]{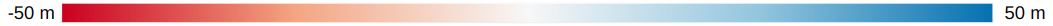}
			\end{minipage}%
			\begin{minipage}[t]{0.35\linewidth}
				\centering
				\includegraphics[width=1\linewidth]{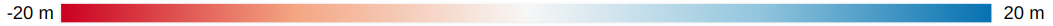}
			\end{minipage}%
		\caption{\textbf{Difference of DSMs.} We compute the differences w.r.t. GT DSMs for the two best performing methods. Although \OurNeRFShort~behaves better near discontinuities in urban DFC dataset, it is unable to recover high frequency details in rural Djibouti. Notice that the difference maps for SGM (g,h) carry a repetitive signal typical for aliasing due to image resampling. Such artefacts are not present in \OurNeRFShort.} 
		\label{DoDComp}
	\end{center}
\end{figure*}

\begin{figure*}[htbp]
    \begin{center}
        \subfigure[Dense Sat-\Nerf~]{
            \begin{minipage}[t]{0.15\linewidth}
                \centering
                    \begin{tikzpicture}
                    \node[anchor=south west,inner sep=0] (image) at (0,0) {
				\includegraphics[width=1\linewidth]{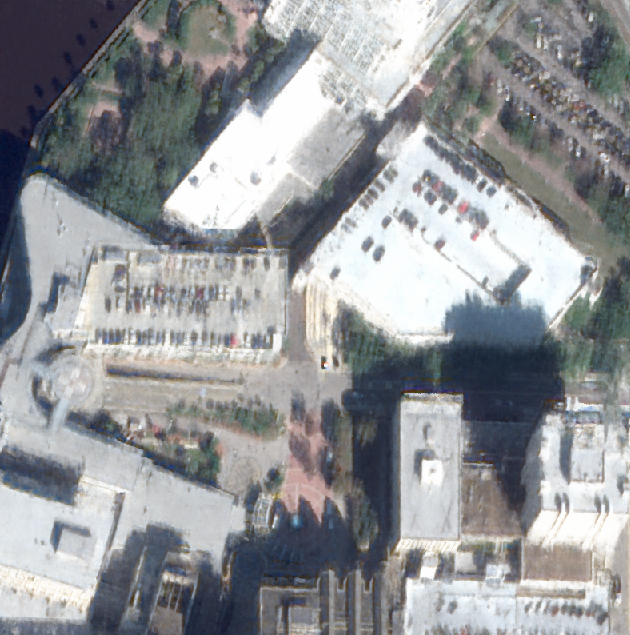}};
                    \begin{scope}[x={(image.south east)},y={(image.north west)}]
                    \draw[red,thick] (0.033,0.326) rectangle (0.171,0.752);
                    \end{scope}
                    \end{tikzpicture}                
            \end{minipage}%
            \vspace{1pt}
			\begin{minipage}[t]{0.052\linewidth}
				\centering
                    \begin{tikzpicture}
                    \node[anchor=south west,inner sep=0] (image) at (0,0) {
				\includegraphics[trim=30 0 100 0,clip,width=1\linewidth]{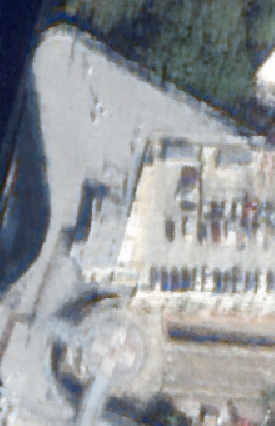}};
                    \begin{scope}[x={(image.south east)},y={(image.north west)}]
                    \draw[red,thick] (0,0) rectangle (1,1);
                    \end{scope}
                    \end{tikzpicture}
			\end{minipage}%
        }
        \subfigure[\OurNeRFShort~$\setminus$Corr]{
            \begin{minipage}[t]{0.15\linewidth}
                \centering
                    \begin{tikzpicture}
                    \node[anchor=south west,inner sep=0] (image) at (0,0) {
				\includegraphics[width=1\linewidth]{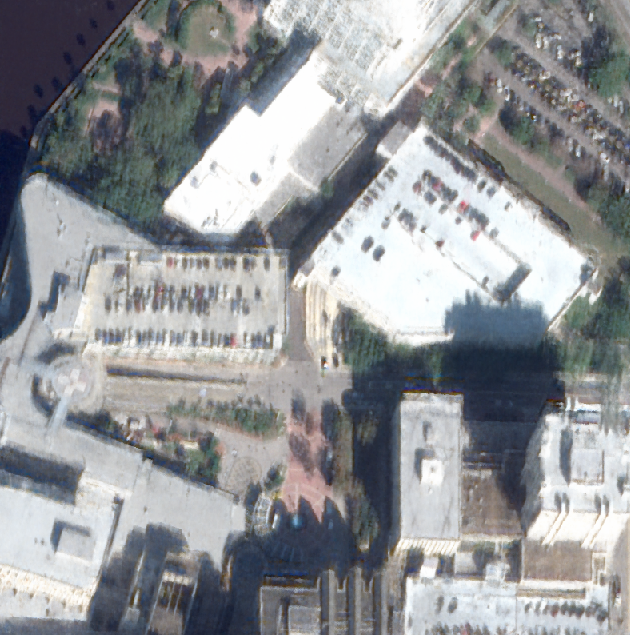}};
                    \begin{scope}[x={(image.south east)},y={(image.north west)}]
                    \draw[red,thick] (0.033,0.326) rectangle (0.171,0.752);
                    \end{scope}
                    \end{tikzpicture}                
            \end{minipage}%
            \vspace{1pt}
			\begin{minipage}[t]{0.052\linewidth}
				\centering
                    \begin{tikzpicture}
                    \node[anchor=south west,inner sep=0] (image) at (0,0) {
				\includegraphics[trim=30 0 100 0,clip,width=1\linewidth]{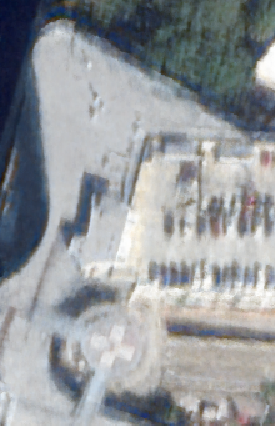}};
                    \begin{scope}[x={(image.south east)},y={(image.north west)}]
                    \draw[red,thick] (0,0) rectangle (1,1);
                    \end{scope}
                    \end{tikzpicture}
			\end{minipage}%
        }
        \subfigure[\OurNeRFShort]{
            \begin{minipage}[t]{0.15\linewidth}
                \centering
                    \begin{tikzpicture}
                    \node[anchor=south west,inner sep=0] (image) at (0,0) {
				\includegraphics[width=1\linewidth]{figures/result/C1-RGB.png}};
                    \begin{scope}[x={(image.south east)},y={(image.north west)}]
                    \draw[red,thick] (0.033,0.326) rectangle (0.171,0.752);
                    \end{scope}
                    \end{tikzpicture}                
            \end{minipage}%
            \vspace{1pt}
			\begin{minipage}[t]{0.052\linewidth}
				\centering
                    \begin{tikzpicture}
                    \node[anchor=south west,inner sep=0] (image) at (0,0) {
				\includegraphics[trim=30 0 100 0,clip,width=1\linewidth]{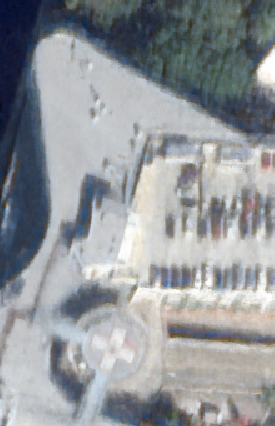}};
                    \begin{scope}[x={(image.south east)},y={(image.north west)}]
                    \draw[red,thick] (0,0) rectangle (1,1);
                    \end{scope}
                    \end{tikzpicture}
			\end{minipage}%
        }
        \subfigure[GT]{
            \begin{minipage}[t]{0.15\linewidth}
                \centering
                    \begin{tikzpicture}
                    \node[anchor=south west,inner sep=0] (image) at (0,0) {
				\includegraphics[width=1\linewidth]{figures/result/GT1-RGB.png}};
                    \begin{scope}[x={(image.south east)},y={(image.north west)}]
                    \draw[red,thick] (0.033,0.326) rectangle (0.171,0.752);
                    \end{scope}
                    \end{tikzpicture}                
            \end{minipage}%
            \vspace{1pt}
			\begin{minipage}[t]{0.052\linewidth}
				\centering
                    \begin{tikzpicture}
                    \node[anchor=south west,inner sep=0] (image) at (0,0) {
				\includegraphics[trim=30 0 100 0,clip,width=1\linewidth]{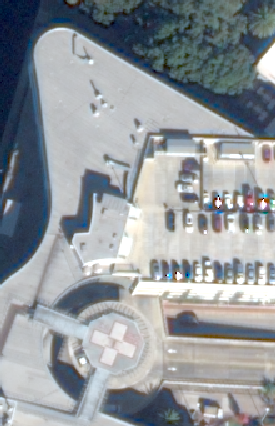}};
                    \begin{scope}[x={(image.south east)},y={(image.north west)}]
                    \draw[red,thick] (0,0) rectangle (1,1);
                    \end{scope}
                    \end{tikzpicture}
			\end{minipage}%
        }\\

        \subfigure[Dense Sat-\Nerf~]{
            \begin{minipage}[t]{0.15\linewidth}
                \centering
                \begin{tikzpicture}
                \node[anchor=south west,inner sep=0] (image) at (0,0) {
                \includegraphics[width=1\linewidth]{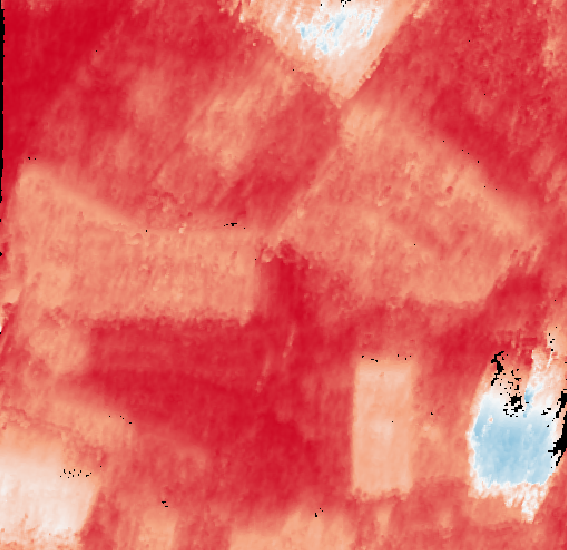}};
                \begin{scope}[x={(image.south east)},y={(image.north west)}]
                \draw[rotate around={-42:(0.8,0.7)}, blue, thick] (0.6,0.65) rectangle (0.982,0.8);
                \end{scope}
                \end{tikzpicture}
            \end{minipage}%
            \vspace{1pt}
			\begin{minipage}[t]{0.052\linewidth}
				\centering
                    \begin{tikzpicture}
                    \node[anchor=south west,inner sep=0] (image) at (0,0) {
				\includegraphics[width=1\linewidth]{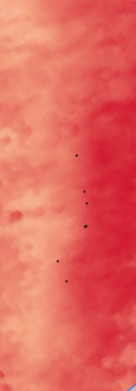}};
                    \begin{scope}[x={(image.south east)},y={(image.north west)}]
                    \draw[blue,thick] (0,0) rectangle (1,1);
                    \end{scope}
                    \end{tikzpicture}
			\end{minipage}%
        }
        \subfigure[\OurNeRFShort~$\setminus$Corr]{
            \begin{minipage}[t]{0.15\linewidth}
                \centering
                \begin{tikzpicture}
                \node[anchor=south west,inner sep=0] (image) at (0,0) {
                \includegraphics[width=1\linewidth]{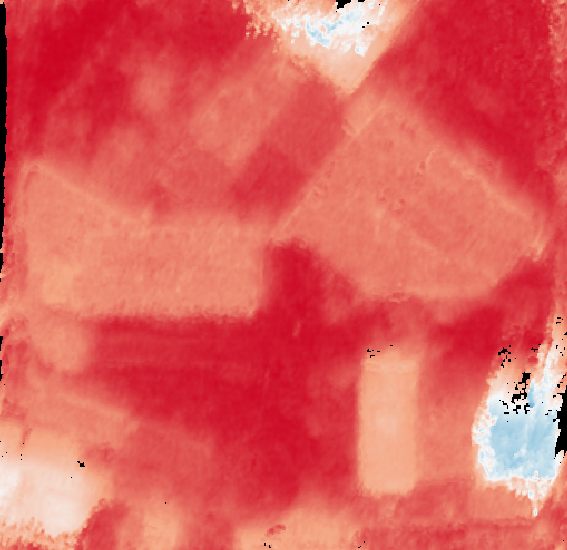}};
                \begin{scope}[x={(image.south east)},y={(image.north west)}]
                \draw[rotate around={-42:(0.8,0.7)}, blue, thick] (0.6,0.65) rectangle (0.982,0.8);
                \end{scope}
                \end{tikzpicture}
            \end{minipage}%
            \vspace{1pt}
			\begin{minipage}[t]{0.052\linewidth}
				\centering
                    \begin{tikzpicture}
                    \node[anchor=south west,inner sep=0] (image) at (0,0) {
				\includegraphics[width=1\linewidth]{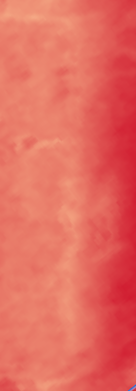}};
                    \begin{scope}[x={(image.south east)},y={(image.north west)}]
                    \draw[blue,thick] (0,0) rectangle (1,1);
                    \end{scope}
                    \end{tikzpicture}
			\end{minipage}%
        }
        \subfigure[\OurNeRFShort]{
            \begin{minipage}[t]{0.15\linewidth}
                \centering
                \begin{tikzpicture}
                \node[anchor=south west,inner sep=0] (image) at (0,0) {
                \includegraphics[width=1\linewidth]{figures/result/C1-DSM.png}};
                \begin{scope}[x={(image.south east)},y={(image.north west)}]
                \draw[rotate around={-42:(0.8,0.7)}, blue, thick] (0.6,0.65) rectangle (0.982,0.8);
                \end{scope}
                \end{tikzpicture}
            \end{minipage}%
            \vspace{1pt}
			\begin{minipage}[t]{0.052\linewidth}
				\centering
                    \begin{tikzpicture}
                    \node[anchor=south west,inner sep=0] (image) at (0,0) {
				\includegraphics[width=1\linewidth]{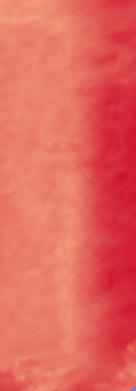}};
                    \begin{scope}[x={(image.south east)},y={(image.north west)}]
                    \draw[blue,thick] (0,0) rectangle (1,1);
                    \end{scope}
                    \end{tikzpicture}
			\end{minipage}%
        }
        \subfigure[GT]{
            \begin{minipage}[t]{0.15\linewidth}
                \centering
                \begin{tikzpicture}
                \node[anchor=south west,inner sep=0] (image) at (0,0) {
                \includegraphics[width=1\linewidth]{figures/result/GT12-DSM.png}};
                \begin{scope}[x={(image.south east)},y={(image.north west)}]
                \draw[rotate around={-42:(0.8,0.7)}, blue, thick] (0.6,0.65) rectangle (0.982,0.8);
                \end{scope}
                \end{tikzpicture}                
            \end{minipage}%
            \vspace{1pt}
			\begin{minipage}[t]{0.052\linewidth}
				\centering
                    \begin{tikzpicture}
                    \node[anchor=south west,inner sep=0] (image) at (0,0) {
				\includegraphics[clip,width=1\linewidth]{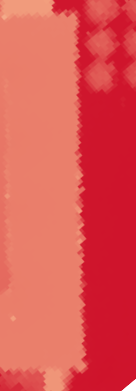}};
                    \begin{scope}[x={(image.south east)},y={(image.north west)}]
                    \draw[blue,thick] (0,0) rectangle (1,1);
                    \end{scope}
                    \end{tikzpicture}
			\end{minipage}%
        }\\
			\begin{minipage}[t]{0.6\linewidth}
				\centering
				\includegraphics[width=1\linewidth]{figures/result/legend12.png}
			\end{minipage}%
        \caption{\textbf{Ablation experiment}. Qualitative result on \Nerf~variants trained with 2 views (DFC2019). {The top row (a-d) represents the novel views, while the bottom row (e-h) shows DSMs.} Adding dense supervision (a,e), guided ray sampling (b,f) and uncertainty measures (c,g) contribute to visually better surface geometries and sharper novel views.} 
        \label{AblationRes}
    \end{center}
\end{figure*}

\subsection{Results \& discussion}

\begin{table*}[htbp]
\scriptsize
\centering
\begin{tabular}{|c|c|c|c|c|}
\hline 
Method & \footnotesize  PSNR $\uparrow$ & \footnotesize SSIM $\uparrow$ & \footnotesize  MAE$_{in}$ $\downarrow$ & \footnotesize MAE$_{out}$ $\downarrow$ \\\hline\hline
\footnotesize Dense Sat-\Nerf & 19.39 & 0.86 & 3.58 & 7.91 \\\hline
\footnotesize \OurNeRFShort~$\setminus$Corr & 19.67 & 0.86 & 3.21 & 8.03 \\\hline
\footnotesize \OurNeRFShort & \textbf{20.2} & \textbf{0.87} & \textbf{3.02} & \textbf{7.77} \\\hline
\end{tabular}
\caption{\textbf{Ablation experiment}.  Quantitative metrics on \Nerf~variants trained with 2 views from DFC2019. Adding dense supervision (Dense Sat-\Nerf), guided ray sampling (\OurNeRFShort~$\setminus$Corr) and uncertainty measures (\OurNeRFShort) improve the novel view generation and surface recovery metrics.}
\label{table:ablation}
\end{table*}

\paragraph{Novel view synthesis}
Qualitative and quantitative results are given in \Cref{RGBComp} and \Cref{PSNR_SSIM_MAE}. In the urban DFC2019 dataset \Nerf's and Sat-\Nerf's novel views are poorly rendered. \OurNeRFShort~provides better quality synthetic views with 2 input images (\Cref{RGBComp}(k)), and further improves the result with 3 input images (\Cref{RGBComp}(l)).
In the rural Djibouti dataset, the performance gap between \Nerf, Sat-\Nerf~and \OurNeRFShort~is less significant, however, in \Cref{RGBComp} \textit{ghost} artifacts are revealed by \Nerf~(c), which are attenuated by Sat-\Nerf~(g) and are not present in \OurNeRFShort~(o).

\paragraph{Altitude extraction}
The qualitative and quantitative results are in \Cref{DSMComp} and \Cref{PSNR_SSIM_MAE}. NeRF fails to recover reasonable DSM geometries for all  4 scenarios (a, b, c, d). This is because using only RGB consistency between input images is insufficient to recover the scene's surface with 2 or 3 images. Adding sparse depth supervision in Sat-\Nerf~helps to recover rough buildings' shapes in $DFC_{3v}$ scenario (f). Nevertheless, it fails at the remaining three scenarios (e, g, h), indicating that sparse depths are not enough to complete the missing information with 2 or 3 input images.\\ 
%
%
Our \OurNeRFShort~takes as input dense depths computed with SGM using downsampled images (factor 4). The input depth maps are incomplete (due to occlusions) and imprecise ($\times4$ bigger GSD), but \OurNeRFShort~is able to complete and refine the depth information. We attribute this to the jointly optimized RGB and depth losses. Compared to the SGM result obtained with full-resolution images ($SGM_{scl1}$), \OurNeRFShort~behaves better close to the outlines of buildings and is free of outliers, but lacks regularization on flat surfaces (see \Cref{DoDComp}). Such local irregularities are a common problem in \Nerf~\cite{mari2022sat}. Adding semantic information to the framework might be a possible solution. Interestingly, \OurNeRFShort~with 3 views is capable of recovering trees' canopy surface (see \Cref{DSMComp} (n)), a task traditionally challenging for traditional patch-based methods such as SGM.\\
%
It should be mentioned that the GT DSM in the Djibouti dataset \Cref{DSMComp}(w, x) was generated with the very same SGM as the best performing $SGM_{scl1}$. This correlation might potentially bias the comparison. Additionally, SGM is susceptible to outliers, as shown in the zoom-in view of GT DSM in \Cref{DSMComp}(w). Hence, our GT DSM is likely corrupt with some erroneous depth estimations.


\paragraph{Ablation study.} We perform two experiments training different variants of \Nerf~with 2 views from the DFC2019 dataset: (i)~\textit{Dense Sat-\Nerf} where we train the vanilla Sat-\Nerf~and replace the sparse depth supervision with our dense depths; (ii) \textit{\OurNeRFShort~$\setminus$Corr} where we train our \OurNeRFShort~and set the $corr(\textbf{r})$=1 for every pixel in \Cref{s_GT} and \Cref{depthloss} thus we deactivate the uncertainty metric but maintain the ray sampling strategy.\\
In \Cref{AblationRes} we compare the novel view and depths generated by \textit{Dense Sat-\Nerf}, \textit{\OurNeRFShort$\setminus$Corr} with our full \OurNeRFShort. Without the guided ray sampling, \textit{Dense Sat-\Nerf} struggles to recover a high contrast image (a) and sharp buildings' outlines (e). The performance  improves in \textit{\OurNeRFShort$\setminus$Corr} (b and f), where the network is encouraged to estimate the depth within the $m$ margin (\Cref{s_GT}) of the input depth while balancing the color loss. The performance is further enhanced by adding $corr(\textbf{r})$ (\Cref{AblationRes}(c, g)). Quantitative results in \Cref{table:ablation} show the same tendencies.



\section{Conclusion}
We presented \OurNeRF~(\OurNeRFShort) -- an extension of Sat-\Nerf~adapted to novel view generation and 3D geometry reconstruction from sparse satellite image views. The adaptation consists of including dense depth supervision with low resolution surfaces obtained with traditional dense image matching, and a suitable ray sampling borrowed from \cite{roessle2022dense}. To add robustness to our supervision we incorporate uncertainty metrics based on dense image matching cross-correlation maps. We demonstrate that \OurNeRFShort~performs better than \Nerf~ and Sat-\Nerf~in sparse view scenarios. It is also competitive with the traditional semi-global matching.

\section{Acknowledgement}
This research was funded by CNES (Centre national d'études spatiales). The Djibouti dataset was obtained through the CNES ISIS framework. The numerical computations were performed on the SCAPAD cluster processing facility at the Institute de Physique du Globe de Paris. We thank Stéphane Jacquemoud and Tri Dung Nguyen for familiarizing us with the Djibouti dataset.

\bibliographystyle{plain} 
\footnotesize
\bibliography{DDSSNeRF}

\end{document}